%% file: main.tex
\pdfoutput=1
\documentclass[11pt, dvipsnames]{article}

\usepackage{EMNLP2023}
\usepackage[shortlabels]{enumitem}
\usepackage{times}
\usepackage{color,soul} 
\usepackage{booktabs}
\usepackage{latexsym}
\usepackage{hyperref}
\usepackage{multirow}
\usepackage[LAE,T1]{fontenc}
\usepackage{arabtex}
\usepackage{utf8} % Needed by arabtex package
\usepackage[utf8]{inputenc} % Needed by LAE, T1 for inline arabic
\usepackage[LAE,T1]{fontenc}
\usepackage[arabic,french,english]{babel}
\usepackage[raggedrightboxes]{ragged2e}
\usepackage{todonotes}
\usepackage{array}
\usepackage{lscape}
\usepackage{natbib}
\usepackage{longtable}
\usepackage{multirow}
\usepackage{float}
\usepackage{makecell}
\usepackage{xcolor, colortbl}
\usepackage{inconsolata}
\usepackage{arydshln}
\usepackage{array}
\usepackage{arydshln}
\newcolumntype{H}{>{\setbox0=\hbox\bgroup}c<{\egroup}@{}}
\usepackage{cancel}
\usepackage{ulem,xpatch}

\newcommand\blfootnote[1]{%
  \begingroup
  \renewcommand\thefootnote{}\footnote{#1}%
  \addtocounter{footnote}{-1}%
  \endgroup
}
\usepackage[T1]{fontenc}
\usepackage[utf8]{inputenc}

\usepackage{microtype}
\usepackage{rotating}

\usepackage{inconsolata}

\usepackage{graphicx}
\usepackage{subcaption}
\usepackage{multirow}
\usepackage{tabularx}
\usepackage{framed}
\title{\includegraphics[scale=0.135]{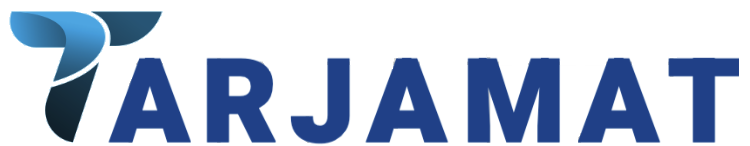}:
Evaluation of Bard  and  ChatGPT on Machine Translation of Ten Arabic Varieties}  

\author{\normalsize Karima Kadaoui$^{\lambda,\star}$~Samar M. Magdy$^{\lambda,\star}$~Abdul Waheed$^{\lambda,\star}$~Md Tawkat Islam Khondaker$^{\xi,\star}~$ \\ 
\normalsize \bf Ahmed Oumar El-Shangiti$^{\lambda}$~El Moatez Billah Nagoudi$^{\xi}$~Muhammad Abdul-Mageed$^{\xi,\lambda,\star}$ \\
\normalsize $^{\xi}$ Deep Learning \& Natural Language Processing Group, The University of British Columbia\\\normalsize $^{\lambda}$Department of Natural Language Processing \& Department of Machine Learning, MBZUAI\\ %
  \texttt{\normalsize \{karima.kadaoui,samar.magdy,abdul.waheed,ahmed.oumar\}@mbzuai.ac.ae}\\
  \texttt{\normalsize \{tawkat@cs,moatez.nagoudi,muhammad.mageed\}@ubc.ca}}

\begin{document}
\setcode{utf8}
\maketitle
% \begin{abstract}
\section*{~~~~~~~~~~~~~~~~~~~~~~~~~~~~~Abstract}
 
Despite the purported multilingual proficiency of instruction-finetuned large language models (LLMs) such as ChatGPT and Bard, the linguistic inclusivity of these models remains insufficiently explored. Considering this constraint, we present a thorough assessment of Bard and ChatGPT (encompassing both GPT-3.5 and GPT-4) regarding their machine translation proficiencies across ten varieties of Arabic. Our evaluation covers diverse Arabic varieties such as Classical Arabic (CA), Modern Standard Arabic (MSA), and several country-level dialectal variants. Our analysis indicates that LLMs may encounter challenges with dialects for which minimal public datasets exist, but on average are better translators of dialects than existing commercial systems. On CA and MSA, instruction-tuned LLMs, however, trail behind commercial systems such as Google Translate. Finally, we undertake a human-centric study to scrutinize the efficacy of the relatively recent model, Bard, in following human instructions during translation tasks. Our analysis reveals a circumscribed capability of Bard in aligning with human instructions in translation contexts. Collectively, our findings underscore that prevailing LLMs remain far from inclusive, with only limited ability to cater for the linguistic and cultural intricacies of diverse communities.\blfootnote{$^\star$Equal contribution}
   
%Experimental setup:
%Dataset: A hundred sentences are selected randomly from OpenITI dataset for CA with a length of not less than ten words each. 
%A hundred sentences are selected randomly from the latest news on the Aljazeera website, along with other news websites. 
%Two hundred sentences are selected from the unreleased DA dataset (for Egyptian and Moroccan dialects) to test ChatGPT's translation ability on the dialectal level.
%A hundred texts are selected from the Parseme dataset for MWEs. 
% \end{abstract}

\input{Sections/intro}
\input{Sections/Lit_Review/abdulLit}

\input{Sections/datasets}
\input{Sections/methodology}
\input{Sections/results}
\input{Sections/bard-helpfulness}

% % I don't think we need this section hence commenting
% \input{Sections/discussion}
 
\input{Sections/conclusion}

% Limitations and Ethics
\input{Sections/limitations}
\input{Sections/ethical-statement}
% \bibliographystyle{plain}
% This specifies the location of the file containing the bibliographic information.  
% It assumes you're using BibTeX (if not, why not?).
\input{Sections/ack}
\clearpage       % Use \clearpage instead if the document class uses the "oneside" argument
\phantomsection  % With hyperref package, enables hyperlinking from the table of contents to bibliography             
% The following statement causes the title "References" to be used for the bibliography section:

\normalem
\bibliography{custom}
\bibliographystyle{acl_natbib}

\appendix
\input{Sections/appendix}

\end{document}

%% file: Sections/intro.tex
\section{Introduction}\label{sec:intro}
Large language models (LLMs) finetuned to follow instructions~\cite{flan_wei, supernatural_wang, instructgpt_ouyang} have recently emerged as powerful systems for handling a wide range of NLP tasks. In accordance with the scaling law (i.e., pretraining larger models will continue to result in better performance)~\cite{kaplan2020scaling}, a number of LLMs such as GPT-3~\cite{gpt_3}, Chinchilla~\cite{chinchilla}, Claude~\cite{claude}, ChatGPT\footnote{In this work, we refer \texttt{gpt-3.5-turbo} as ChatGPT.}~\cite{chatgpt}, GPT-4~\cite{gpt_4}, and Bard~\cite{bard} have been introduced. Most of these models, however, are `closed'. That is, little-to-no information about them is known. This includes details about model architectures, pretraining data, languages involved, and training configurations. LLMs are also expensive both to pretrain and deploy. To alleviate these concerns, `open' LLMs such as BLOOM~\cite{bloom}, LLaMA-1~\cite{llama_1}, Falcon~\cite{falcon}, and LLaMA-2~\cite{llama_2} were introduced. These more open models can facilitate research and (non-) commercial deployment.

\begin{figure}[t]
    \centering
    \includegraphics[width = \columnwidth]{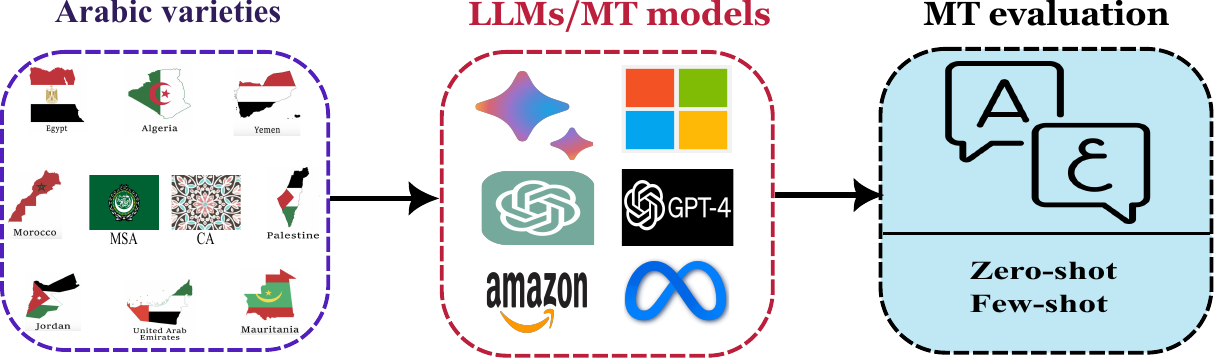}
    \caption{Experimental setup for our evaluation. We evaluate multiple language models on different Arabic varieties.}%\protect\footnotemark}
    \label{fig:pipeline}
    % \protect\footnotetext{A larger version of this image in our Appendix.}
    % \protect\footnotetext{A larger version of this image in our Appendix.}

\end{figure}
%\protect\footnotetext{A larger version of the pipeline diagram is in Appendix~\ref{appendi:setup}.}

In spite of drawbacks such as their closed nature, computational costs~\cite{distillation-2023-cost}, and biases they exhibit~\cite{ferrara2023should}, closed LLMs remain attractive primarily due to their remarkable performance~\cite{Bang2023AMM, laskar2023systematic}. It is thus important to fully understand the full capabilities of these closed models. Although there has been a recent flurry of works attempting to evaluate ability of LLMs to carry out NLP tasks, many of these models remain opaque. This is especially the case when it comes to understanding how LLMs fare on different varieties and dialects of several popular languages and on vital tasks such as machine translation (MT). For example, the extent to which LLMs can handle MT from Arabic varieties into other languages is unknown. 

Another challenge is how more recent models such as Google's Bard are yet to be evaluated and understood. Bard was released in $41$ different languages, which makes it a particularly attractive target for MT evaluation. This is also the case given Google's strong history of investment in MT~\cite{Wu2016GooglesNM}. In this work, we offer a thorough evaluation of LLMs on MT from major Arabic varieties into English (Figure~\ref{fig:pipeline}). Namely, we evaluate ChatGPT, GPT-4, and Bard on MT of ten Arabic varieties into English. Since there are usually concerns about downstream evaluation data leaking into LLM pretraining, which involves data collected from the web, we benchmark the models on new test sets that we manually prepare for this work. Our evaluation targets diverse varieties of Arabic. Namely, we evaluate on Classical Arabic (CA), Modern Standard Arabic (MSA), and several country-level Arabic dialects such as Algerian and Egyptian Arabic (Section~\ref{sec:data}).% \textcolor{red}{(see Section xx)}. 

Bard provides three different drafts for each text input we ask it to translate. Contents of the three drafts are diverse, providing us with excellent contexts to analyze the degree to which the model adheres to our prompts. We leverage these contexts to carry out a human evaluation study investigating the \textit{helpfulness} of the model, allowing us to reveal a number of Bard's limitations. We carefully analyze these limitations against the different Arabic varieties we target, thus affording even better understanding of the model's ability to translate from Arabic. 

Overall, our work offers the following contributions:  
\begin{enumerate}[(i)]
    \item We offer a detailed MT evaluation of instruction finetuned LLMs on ten diverse varieties of Arabic.
    \item To the best of our knowledge, our work is the first to assess performance of Bard on NLP tasks in any language, and on Arabic MT in particular.
    \item We introduce a new manually created multi-Arabic dataset for MT evaluation that has never been exposed to any existing LLM.
    \item We extensively evaluate Bard through a human study to analyze its behavior in terms of \textit{helpfulness}. We examine how well the model follows human instructions when tasked with translating across ten different Arabic varieties.
\end{enumerate}

The rest of the paper is organized as follows:  In Section~\ref{sec:rel}, we review previous research evaluating LLMs on NLP tasks in general and MT in particular. % We also delve into the MT capabilities of ChatGPT, focusing on Arabic MT for both Modern Standard Arabic (MSA) and dialectal language. Additionally, we review related work on ChatGPT's application to Arabic MT translation. 
In Section~\ref{sec:data}, we introduce our newly developed multi-Arabic MT dataset. In Section~\ref{sec:methodology}, we describe our evaluation methods. %This includes the design of the initial prompts, a pilot experiment, and the assessment of the prompts. Furthermore, we conduct N-Shot experiments, exploring the performance of the models in various settings such as 0-shot, 1-shot, 3-shot, and 5-shot. We also explain the evaluation metrics and baselines employed to assess the performance of ChatGPT and Bard. 
In Section~\ref{sec:results}, we present our results and the main findings obtained from comparing ChatGPT and Bard to various commercial MT products. In Section~\ref{sec:human_analysis_of_bard_helpfulness}, we present our human study analyzing Bard's helpfulness, particularly in terms of its ability to follow human instructions in MT. We conclude in Section~\ref{sec:conclusion}. %, we summarize our findings and propose suggestions for further research. 

%% file: Sections/Lit_Review/abdulLit.tex
% \textcolor{red}{[Language needs to be fixed!]}
% \moa{Can we plz add a short paragraph here as an introduction}

\section{Related Work}\label{sec:rel}

%In this section, we describe and critique various works related to Arabic MT and evaluation of ChatGPT, GPT-4, and Bard for various NLP tasks with a focus on MT.

\noindent\noindent\textbf{Evaluation of ChatGPT and Other LLMs.}
A growing body of literature has focused on evaluating ChatGPT and other LLMs on NLP tasks. \citet{laskar2023systematic} find ChatGPT effective on many tasks. Other works find it either on par with supervised models \cite{ziems-2023-can} or in some cases (e.g., sequence tagging) falling behind these  models~\cite{qin-2023-chatgpt}. Both \citet{jiao2023chatgpt} and~\citet{ogundare2023comparative} find that GPT-4 is competitive with commercial systems for high-resource languages but lags behind for low-resource languages. \citet{bang2023multitask} find a similar pattern for ChatGPT.  \citet{guerreiro2023hallucinations} find complex translation scenarios, such as in the low-resource setting, to be prone to hallucination. \citet{peng2023making} demonstrate that ChatGPT can surpass Google Translate on many translation pairs, but \citet{zhu2023multilingual} show it is outperformed by NLLB~\cite{nllbteam2022language} on at least $83$\% of the English-centric pairs they study. \citet{wang2023documentlevel,karpinska2023large}, however, show that ChatGPT can match the performance of fully supervised models for document-level translation. 

\citet{peng2023making} find that adding task and domain-specific information in the prompt can improve the robustness of the MT system, which corroborates the findings by \citet{gao2023design}. \citet{huang2023languages} propose a prompting technique called cross-lingual-thought prompting (XLT) to improve cross-lingual performance for a wide range of tasks, including MT. Similarly, \citet{lu2023error} asks ChatGPT to correct its own mistakes as a way to improve the model's translation quality. \citet{lu2023chainofdictionary} propose Chain-of-Dictionary (CoD) prompting to solve rare word translation issues. Prompting with CoD improves the performance of ChatGPT for both X-En and En-X language directions.

\noindent\textbf{Evaluation of ChatGPT on Arabic.}
%\citet{khondaker2023gptaraeval} and \citet{alyafeai2023taqyim} evaluate ChatGPT for X-Arabic and Arabic-X translation pairs. \
\citet{khondaker2023gptaraeval} evaluate ChatGPT and other contemporary LLMs such as BloomZ~\cite{muennighoff2022crosslingual} in few-shot settings (0, 1, 3, 5, and 10) on four X-Arabic and two code-mixed Arabic-X language sets. They show that providing in-context examples to ChatGPT achieves comparable results to a supervised baseline. \citet{alyafeai2023taqyim} evaluate ChatGPT and GPT-4 on $4,000$ Arabic-English sentence pairs from~\citet{ziemski2016united} and find a supervised SoTA model to outperform ChatGPT and GPT-4 by a significant margin. These works, however, only consider a limited number of Arabic varieties. They also do not conduct a thorough analysis of the LLMs for MT. Additionally, none of the works evaluate Bard. Our work bridges these gaps by performing a comprehensive evaluation of these systems on a wide range of Arabic varieties. We also conduct our study on novel in-house data that we guarantee no leakage for (i.e., our data cannot have been seen by ChatGPT, GPT-4, or Bard since we create the data for this work).  
% Our work is unique in evaluating on  a wider range of dialects and we offer a human study of Bard usefulness. We now introduce our datasets.
Other works have focused on evaluating smaller-sized Arabic language models~\cite{abu-farha-magdy-2021-benchmarking,inoue2021interplay,app12115720}, including on recent benchmarks~\cite{nagoudi2023dolphin, elmadany2022orca}. %% Add all related works.
% Arabic has a plethora of dialects, which is one reason that makes Arabic languages very complex

\noindent\textbf{Arabic MT.} There are several works on Arabic MT itself, including rule-based ~\cite{bakr2008hybrid,mohamed2012transforming,salloum2013dialectal}, statistical ~\cite{habash2009improving,salloum2011dialectal,ghoneim2013multiword}, and neural~\cite{junczys2016neural,almahairi2016first,durrani2017qcri,alrajeh2018recipe}. While these systems focus on MSA, others target Arabic dialects~\cite{zbib2012machine,sajjad2013translating,salloum2014sentence,guellil2017neural,baniata2018neural,sajjad2020arabench,farhan2020unsupervised,nagoudi-2021-Code-Mixed,nagoudi2022_arat5}. We provide a more detailed review of related literature in Appendix \ref{sec:appendix-rel}, with a summary in Table~\ref{literature-review-table}.

%% file: Sections/datasets.tex
\section{Coverage and Datasets}\label{sec:data}

%%%%%%%%%%%%%%%%%%%%%%%%%%%%%%%%%%%%%%%%%%%%%%%%%%%%%%%%%%%%%%%%%%%%%
%\subsection{Translation Datasets}\label{subsec:transData}

% As explained earlier, we manually develop test sets for ten different varieties of Arabic. We provide data from \textbf{CA because it has distinct properties such as xxx}. 

% In this report, 
\subsection{Arabic Varieties}
Our goal is to provide a comprehensive evaluation of MT on ChatGPT, GPT-4, and Google Bard, focusing on their performance across ten different varieties of Arabic. These can vary across \textit{time} (i.e., old vs. modern day) and \textit{space} (e.g., country-level geography) as well as their \textit{sociopragmatic} functions (e.g., standard use in government communication vs. everyday street language). Before introducing our dataset, we provide a brief background about Arabic and its varieties.
\noindent Arabic, the collection of languages spoken by approximately $450$ million people across the Arab world, encompasses a broad spectrum of varieties. Classical Arabic (CA) is known as Quranic Arabic, the language of the Quran~\cite{beginnings-classical-arabic}, and has emerged from the medieval dialects of the Arab tribes. It was spoken early in Mecca around $1,500$ years ago in the sixth or seventh century AD. CA is considered the most eloquent form of Arabic and is preserved notably in the Holy Quran and pre-Islamic epic poems \cite{versteegh2014arabic}. It is often described as exhibiting archaic words, figurative speech, and rhyming sentences that are no longer (or less frequently) used in MSA and dialectal Arabic varieties. 
Modern Standard Arabic (MSA)~\cite{holes2004modern}, on the contrary, is deeply rooted in CA that has been lightened to a great extent to encompass the modern uses in Modern literature, poetry and official statements. MSA additionally serves as the standardized language for formal events, news broadcasts, sermons, and formal communication. We now explain how we acquire our dataset for each Arabic variety. %The selected sentences demonstrate the CA's distinct linguistic features regarding word richness, semantic variations, and the usage of Tashkeel.

%mean, median, mode 
% \input{emnlp2023-latex/Tables/stat_sent_dia}
\input{Tables/MT_Examples}

\input{Tables/prompet_table}

\subsection{Datasets}

\noindent \textbf{CA.} %We collect a total of $200$ sentences from the OpenITI~\cite{nigst2020openiti} dataset. The sentences are drawn from various books that belong to the second century. \textcolor{purple}{\small\textbf{(K)} Maybe remove prev 2 sentences, the ones after make them redundant} 
We manually curate $200$ sentences from the Open Islamic Texts Initiative (OpenITI)~\cite{nigst2020openiti} dataset, namely from the latest 2022.16 version. It includes a collection of premodern Arabic works featuring a comprehensive library of $10,342$ books. The sentences were chosen based on a set of specified criteria:
 Initially, we identify books originating from the first and second-century Anno Hegirae (in the year of the Hijra), excluding those written after this period. Then we compile a collection of $15$ distinctive books, including notable works like Abdullah Ibn AlMuqfaa’s ``Al-Adab Al-Kabir” and ``Al-Adab Al-Saghir”, Mohamed Idis Al-Shafi's ``Al-Umm”, ``Al-Risala”, and ``Al-Adab Wal-Muraa”, among others. We subsequently extract sentences of a minimum of ten words. We provide the list of the $15$ books we sample from in Appendix~\ref{appendix-datasets} (Table~\ref{tab:openiti-books}). 
 % These books are: 
 % {\begin{arabtext} 
     
 % \end{arabtext}
\noindent \textbf{MSA.} We collect a total of $200$ sentences from current event news picked from two online news websites:
    % 50 from each of 
    \href{https://aljazeera.net/news}{Aljazeera}\footnote{\url{https://aljazeera.net/news}} and \href{https://bbc.com/arabic}{BBC Arabic}\footnote{\url{https://bbc.com/arabic}}. The curated sentences showcase various news genres, including political, social, and sports.
    % The length of the sentence is not less than ten words each.

\noindent \textbf{Various Dialects.} We manually select a dataset of dialectal Arabic from an in-house project where we transcribe TV series collected from YouTube videos belonging to Arabic dialects. Again, we use $200$ sentences from each dialect, resulting in a total of $1,600$ sentences across eight dialects, each transcribed and translated by their respective native speakers. The dialects belong to North African countries such as Algeria, Morocco, and Mauritania; Gulf area dialects, namely Emirati;  Levantine Arabic (focusing on Palestinian and Jordanian); and Egyptian Arabic.
    % \item \textbf{MWEs:} We collected 100 sentences of verbal MWEs in Modern Standard Arabic from \cite{mohamed2022annotating} the PARSEME dataset.
    % \item \textbf{Syntactic Ambiguity:} One of the linguistic features we are testing is ChatGPT's ability to translate syntactic ambiguous sentences in Arabic. We collected 40 MSA ambiguous sentences from different theses tackling this feature in the Arabic language. 
    % \item \textbf{Collocations:} 
    % \item \textbf{Proverbs:} We selected a hundred proverbs in MSA from websites teaching Arabic as a foreign language. 
    % 150 Egyptian proverbs are collected from websites teaching Egyptian Arabic to non-natives and their meanings in English. 
    % \todo{add MOR}
    % Examples of these proverbs are provided in Table \ref{tab:proverb_examples}.
    % \item \textbf{Synonyms and Antonyms:} To test ChatGPT's linguistics abilities in understanding synonyms and antonyms, 50 words in Egyptian and Moroccan dialects are selected from \cite{abdul2020dialex} Dialex dataset for dialectal Arabic word embedding and the Darija Open Dataset (DODa) respectively \cite{outchakoucht2021moroccan} \todo{Review section after word selection is done}. 
    % \item \textbf{Hypernyms and Hyponyms} In order to test ChatGPT's lexical capabilities in Egyptian and Moroccan dialects, we curated 50 words from \cite{abdul2020dialex}. When selecting nouns, we tend to choose words that are not code-switched or borrowed and incorporate a wide range of hyponyms. 

\noindent For all varieties, we collect sentences that are \textit{at least ten words} long. We present one sample from some of the dataset in Table \ref{tab:variety_examples}. Statistics of the datasets across the Arabic varieties is presented in Appendix~\ref{appendix-datasets} (Table~\ref{tab:length_stats}).

% \end{landscape}

% One of the linguistic features we are testing is ChatGPT's ability to translate syntactic ambiguous sentences in Arabic. We collected 40 MSA ambiguous sentences from different theses tackling this feature in the Arabic language. 

% \begin{table*}[t]
%     \centering
%     \begin{tabular}{m{0.26\linewidth}m{0.26\linewidth}m{0.26\linewidth}}
%     \toprule
%     \Centering{\textbf{Example}} & \Centering{\textbf{Meaning 1}} & \Centering{\textbf{Meaning @} } \\
%     \midrule
%     \RaggedRight{\begin{arabtext} \small جاء الآباء والأبناء المتميزون \end{arabtext}} & The distinguished fathers and sons came & The fathers and the distinguished sons came\\ 
%     \hline
%     \RaggedRight{\begin{arabtext} \small قال أحمد في المسجد يصلي الناس. \end{arabtext}} & Ahmed said that people pray in the mosque & Ahmed said in the mosque that people pray\\ 
%     \hline
%     \RaggedRight{\begin{arabtext} \small انقاذ أهرامات مصر القديمة \end{arabtext}} & The rescue of the ancient pyramids of Egypt & The rescue of the pyramids of ancient Egypt\\ 
%     \bottomrule
%     \end{tabular}
%     \caption{Examples from the syntactic ambiguity dataset with their two possible English translations.}
%     \label{tab:synAmb_examples}
% \end{table*}

%%%%%%%%%%%%%%%%%%%%%%%%%%%%%%%%%%%%%%%%%%%%%%%%%%%%%%%%%%%%%%%%%%%%%

%% file: Tables/MT_Examples.tex
\begin{table}[t]
\centering
\footnotesize 
 \renewcommand{\arraystretch}{1.4}
\resizebox{0.95\columnwidth}{!}{%
\begin{tabular}{cc}
\toprule
\textbf{Variety}                                           & \textbf{Example with English Translation}  \\ \midrule
\rowcolor{blue!5} & \begin{tabular}[c]{@{}r@{}}\<\small ماحنا لو فضلنا مدارين و مستخبين هنموت من الخوف.>\end{tabular}\\ 
\rowcolor{blue!5}  \multirow{-2}{*}{\textbf{EGY}} & \begin{tabular}[c]{@{}l@{}}
And if we keep hiding, we’re going to die out of
fear\end{tabular}\\

\multirow{2}{*}{\textbf{JOR}} & \begin{tabular}[c]{@{}r@{}}\<\small أنا مش مستخف فيه و لا يمكن استخف فيه مهما كان بضل ابوي>\end{tabular}\\   
& \begin{tabular}[c]{@{}l@{}}
I do not and cannot underestimate him; he is still
my father, no matter what.\end{tabular}\\

\rowcolor{blue!5} & \begin{tabular}[c]{@{}r@{}}\<\small    شوف آن شي كامل ادخلتو ماتيت انركيلي فيه خالك اللا القدام.>\end{tabular}\\ 
\rowcolor{blue!5} \multirow{-2}{*}{\textbf{MAU}} & \begin{tabular}[c]{@{}l@{}}
Look, whenever I'm in, I never take a step back; I only go forward.\end{tabular}\\ 

\multirow{2}{*}{\textbf{YEM}} & \begin{tabular}[c]{@{}r@{}}\<\small ركزت لي نفطة تفتيش، في الباب، بتفتش ذي داخلي و ذي خارجي.>\end{tabular}\\   
& \begin{tabular}[c]{@{}l@{}}
I set up a checkpoint at the door to screen anyone who comes in or out.\end{tabular}\\ 

\bottomrule
\end{tabular}
}
\caption{Example sentences from some of the Arabic varieties in our new translation evaluation dataset. See Appendix Table~\ref{output-samples} for remaining varieties.}
    \label{tab:variety_examples}
\end{table}

%% file: Tables/prompet_table.tex
\begin{table}[t]
\centering
\footnotesize 
 \renewcommand{\arraystretch}{1.4}
\resizebox{0.95\columnwidth}{!}{%
\begin{tabular}{clr}
\toprule
\textbf{Prompt}                                           & \multicolumn{1}{c}{\textbf{Template}}                                                                                                                                                                                                                                                                                                                                                   & \textbf{BLEU}  \\ \midrule
\textbf{ENG}                                                       & \begin{tabular}[c]{@{}l@{}}Translate the following Modern Standard Arabic (MSA)\\ sentence into English\end{tabular}                                                                                                                                                                                                                                                                   & \textbf{$48.48$} \\ \hline
\rowcolor{blue!5} \textbf{MSA}                                                        &         
% \begin{tabular}[l]{@{}l@{}}\< ترجم الجملة العربية الفصحى العصرية التالية اللي اللغة الإنجليزية >\end{tabular}                                                                                    
\begin{tabular}[l]{@{}l@{}}

\<  ترجم الجملة العربية الفصحى العصرية التالية  الي اللغة الإنجليزية > \\
% \< >

\end{tabular}   
                                                                                         
& $47.92$          \\ \hline
\begin{tabular}[c]{@{}c@{}} \textbf{ENG} \\ (\textbf{elaborate})\end{tabular} & \begin{tabular}[c]{@{}l@{}}I want you to act as an expert translator. You will translate\\ Modern Standard Arabic (MSA) sentences into English. \\ I will give you a Modern Standard  Arabic (MSA) input,\\ and you will translate it into English and keep the same \\semantic meaning. Please translate this Modern\\ Standard Arabic (MSA) text into English\end{tabular} & $46.17$          \\ \bottomrule
\end{tabular}
}
\caption{
\label{tab:prompt_design_result}
Performance of ChatGPT on the MSA$\rightarrow$English translation task. Our concise English prompt outperforms other prompts in BLEU score.
}
\end{table}

%% file: Sections/methodology.tex
% \input{emnlp2023-latex/Tables/prompet_table}

\section{Methodology}\label{sec:methodology}

\subsection{Prompt Design}

\input{Sections/prompt_design}

\subsection{\textit{N}-Shot Experiments}

We run ChatGPT MT generation under $0$-shot, $1$-shot, $3$-shot, and $5$-shot settings. For a particular translation task, we always select the samples for these in-context learning experiments from the same set of training examples. This means that for a $k$-shot setting, we make sure that if a training sample is selected then it will also be selected for $n$-shot settings where $n$ $>$ $k$. We generate translation with ChatGPT (\texttt{gpt-3.5-turbo}\footnote{Snapshot of \texttt{gpt-3.5-turbo} from June 13th 2023.}, an optimized version of GPT-3.5 series) by setting the temperature to $0.0$ to ensure \textit{deterministic and reproducible results}. In addition, we restrict the maximum token length to $512$ for all the generation tasks. For GPT-4, we use the web interface for MT generation under $0$-shot and $5$-shot settings. For Bard\footnote{Update from - 2023.07.13}, we use the web interface but opt out of generating any few-shot response because it lacks an API and its outputs can be problematic requiring intensive manual preprocessing (Section~\ref{sec:human_analysis_of_bard_helpfulness}).
%We also manually evaluate \textcolor{purple}{\small\textbf{(K)} Manually evaluate? Also may be redundant with the newly added blue paragraph} ChatGPT using the \textit{GPT-4 engine} and the Bard model under the \textit{0-}shot condition only using the web interface of each model.\footnote{In the future, we plan to add results of experiments under the same \textit{N}-shot conditions from the set \textit{\{1,3,5\}} similar to our \texttt{gpt-3.5-turbo} experiments.} 

 \subsection{Evaluation and Baselines}
 % \textcolor{red}{AbdulWaheed}
 % add a citation for metrics 
% metrics that we are going to use for this teration
\newcommand{\metrics}{BLEU~\cite{bleu}, COMET~\cite{rei-etal-2020-comet}, ChrF~\cite{chrf}, ChrF++, and TER~\cite{snover-etal-2006-study}}
\noindent\textbf{Evaluation metrics.} Different evaluation metrics are usually employed to automatically evaluate MT systems. These metrics are often based on word overlap and/or context similarity between references and model outputs. In our work, we employ both types of metrics to evaluate the quality of various translation systems that we consider in our study. Namely, we use \metrics.~We provide a detailed description of each metric in Appendix \ref{appendix-evaluation-metrics}.

%  % moved to appendix
% \noindent\textbf{\textit{BLUE.}}~\cite{bleu} is used to evaluate machine translation quality by comparing n-gram ($n=4$) overlap between machine-generated translations and human references. Higher scores indicate better translation quality.\\
% \noindent\textbf{\textit{COMET.}}~\cite{comet} Cross-lingual Opus METric measures translation quality through source-to-translation word-level alignment. Higher values indicate better quality.\\
% \noindent\textbf{\textit{ChrF and ChrF++.}}.~\cite{chrf} Character n-gram F-score calculates the F-score of character n-grams in the machine translation compared to the reference translations, with higher scores denoting better quality. ChrF++ is an extension of ChrF where the word order is 2.\\
% \textbf{\textit{TER.}}\cite{snover-etal-2006-study} Translation Error Rate measures translation quality by counting edit operations between the machine and reference translations, providing a lower score for better quality.\\
% We use huggingface's implementation of these metrics in \textit{evaluate}\footnote{https://github.com/huggingface/evaluate} package. We use all the default parameters unless otherwise specified above.  

% \noindent{\textbf{Baselines}.} 
\noindent\textbf{Baselines.}
We compare instruction-tuned LLMs to a number of MT systems, including both commercial services (Amazon, Google, and Microsoft) as well as the supervised NLLB-200 system~\cite{nllbteam2022language}\footnote{For NLLB-200, we use the distilled 1.3B}. We provide more details about each of these systems in Appendix~\ref{appendix-baselines}.

\input{Tables/error_analysis}

%% file: Sections/prompt_design.tex
\noindent The term \textit{prompt} refers to the set of instructions used to program an LLM with a goal to steer and enhance its purpose and capabilities~\citep{white_prompt}. Prompts can influence subsequent interactions with the model as well as its generated outputs. Therefore, it is important to clearly identify the right prompts to obtain the desired outcome for a particular task. To determine the right prompt for our translation task, we set up a pilot experiment that we now describe.

\noindent \textbf{Pilot experiment.}  In our pilot experiment, we investigate three prompt candidates. To limit the search space, we perform this experiment only with ChatGPT. We experiment with both Arabic and English prompts to \textit{concisely} instruct ChatGPT to translate from an Arabic variety into English, again restricting our search space to MSA as a variety that is known to overlap with other varieties at all linguistic levels~\cite{abdul-mageed-etal-2020-toward,habash2022introduction}. We also experiment with an \textit{elaborate} English prompt that clearly defines the role and the objective of ChatGPT before asking the model to carry out the translation task. We then evaluate the performance of ChatGPT on $100$ MSA$\rightarrow$English samples. We present the prompt templates and the corresponding performance we acquire in Table~\ref{tab:prompt_design_result}.

\noindent \textbf{Evaluation.} As evident, the concise English prompt outperforms the other two prompts, including the Arabic counterpart (by 1$\sim$2 BLEU scores). This result substantiates findings in prior works~\citep{khondaker2023gptaraeval,lai2023chatgpt} regarding the superiority of English prompts on ChatGPT over non-English prompts. Therefore, in the rest of the paper we employ the concise and direct English prompt to conduct our experiments.

% Various studies have examined ChatGPT's proficiency as a translation instrument, with many emphasizing the crucial role of prompts in optimizing translation outcomes. 
 
% We employed previously designed prompts while also modifying some of them to meet our translation objectives effectively, particularly regarding dialectal Arabic. We chose prompts across diverse domains, which aid ChatGPT in accurately identifying and translating the texts. Using prompts, including instructions to translate from the source to the target language, has proven to be a successful strategy in obtaining high-quality translations from ChatGPT.

%  The initial assessment in our evaluation process involves examining the translation abilities of ChatGPT. Consequently, we used prompts that encompass the domain related to the specific variety under scrutiny to achieve optimal performance. 

%  We queried ChatGPT with the prompt, "\textit{Translate the following [Arabic Variety] sentence into English: [Sentence]}" employing both zero and five-shot approaches.
 % We employed both zero-shot and five-shot prompts to evaluate Dialectal Arabic (DA) to ensure enhanced performance.

%% file: Sections/results.tex
\section{Results and Discussion}\label{sec:results}
\newcommand{\zeroshot}{0-shot}
\newcommand{\oneshot}{1-shot}
\newcommand{\threeshot}{3-shot}
\newcommand{\fiveshot}{5-shot}
\newcommand{\ca}{CA}
\newcommand{\msa}{MSA}
\newcommand{\alg}{ALG}
\newcommand{\egy}{EGY}
\newcommand{\jor}{JOR}
\newcommand{\maur}{MAUR}
\newcommand{\mor}{MOR}
\newcommand{\pal}{PAL}
\newcommand{\uae}{UAE}
\newcommand{\yem}{YEM}

% Define how you want to write metrics and model names
\newcommand{\bleu}{BLEU}
\newcommand{\comet}{COMET}
\newcommand{\chrfpp}{ChrF++}
\newcommand{\gptfour}{GPT-4}
\newcommand{\chatgpt}{ChatGPT}
\newcommand{\msft}{Microsoft}
\newcommand{\amzn}{Amazon}
\newcommand{\gt}{Google}

\newcommand{\metricsmain}{BLEU, COMET, and ChrF++}
We evaluate all models on X-English translation direction where X is an Arabic variety (MSA and CA). As mentioned earlier, we evaluate LLMs (ChatGPT, \gptfour, and Bard) in \textit{n}-shot settings. We report \metricsmain~ in Table \ref{results-main}. We report additional metrics in Appendix~\ref{appendix-results}. We summarize our main findings here.

% \input{Tables/results-tables/bleu-main}
% \input{Tables/results-tables/chrf++-main}
% \input{Tables/results-tables/comet-main}
\input{Tables/results-tables/results-main}
\noindent\textbf{Is \gptfour~ better than \chatgpt?} \textit{In most cases, yes}. GPT-4  consistently outperforms  ChatGPT on many dialects and varieties. However, for JOR and UAE, \chatgpt~ \zeroshot~ performs better than \zeroshot~\gptfour. Overall, on average, \gptfour~ \zeroshot~ outperforms ChatGPT \zeroshot~ by $1\sim3$  points on all metrics. Additionally, GPT4 in \zeroshot~ setting is on par with ChatGPT in the 5-shot setting.
\noindent When comparing ChatGPT with GPT-4 under $5$-shot setting, we observe that ChatGPT substantially closes the performance gap, even outperforming GPT-4 in $6$ out of $10$ varieties in terms of BLEU score. Although GPT-4 marginally outperforms ChatGPT on average BLEU score, \textit{this result shows that by providing few-shot examples, it is possible for ChatGPT to achieve comparable performance to GPT-4 on Arabic MT.}

% Add 0 and 5 shot comparisons - update from previously which was only for zero shot. 
%  Better in zero-shot but comparable in 5-shot -> gap decreases with more shots. 

\noindent\textbf{Is ChatGPT/GPT4 better than Bard?} \textit{In most cases, yes}. For fairness, we compare Bard, ChatGPT, and GPT-4 only under the \zeroshot~condition. In the majority of the varieties, either ChatGPT or GPT-4 outperforms the best Bard draft (i.e., Draft 1). Our results show that Bard is better than both of these models in only three cases (i.e., CA, EGY and JOR). Overall, GPT-4 ranks best (\bleu~score at $23.12$), followed by ChatGPT ($21.77$ \bleu~points), which in turn is followed by Bard ($20.47$ \bleu~points). 
%When trained in-context with any number of shots where $n > 0$, however, ChatGPT outperforms both GPT-4 and Bard (again, the latter two only evaluated under 0-shot in this work).
%Conclusion remains same but numbers have changed

%\textcolor{red}{Section About 0-shot Vs Few-shot}

\noindent\textbf{Is ChatGPT/GPT4 better than commercial systems?} \textit{Yes, but only on dialects}.
We evaluate three commercial translation systems, namely, Amazon, Microsoft, and Google Translate. Among commercial systems, we find Google Translate to outperform other commercial systems across all varieties except YEM. The average score for Google Translate is $22.29/64.89/43.11$ (\bleu/\comet/\chrfpp) compared to $18.80/63.68/41.55$ and  $17.77/62.85/39.76$ for Microsoft and Amazon systems, respectively.

From our evaluation results in Table~\ref{results-main}, we observe that commercial systems are better at translating CA and MSA but fail to produce high-quality translations when it comes to dialectal Arabic. ChatGPT and \gptfour~in $0$-shot and few-shot settings are on par or better than the best-performing commercial system (i.e., Google Translate) for all Arabic dialects except JOR. The average \bleu~score of ChatGPT and \gptfour~ in few-shot setting is $23.62$ ($5$-shot) and $13.64$ ($5$-shot), respectively, compared to $2.29$ for Google Translate. However, we notice that Google Translate outperforms ChatGPT and GPT-4 on MSA by a significant margin (while it stays behind on other dialects). Hence, we conclude that \textit{ChatGPT and GPT-4 are better translators of Arabic dialects than the commercial Google Translate system}. We find similar patterns in other metrics. 

\noindent\textbf{Is ChatGPT/\gptfour~ better than the supervised baseline?} \textit{Yes, it is}.
We evaluate NLLB~\cite{nllbteam2022language} as the supervised baseline, finding both ChatGPT and \gptfour~able to outperform this baseline in the \zeroshot~setting. The average \bleu~score for NLLB is $12.97$ compared to $21.77$ and $23.12$ of ChatGPT and GPT-4 under $0$-shot settings, respectively. Similar to the commercial systems, the supervised baseline (NLLB) does well on MSA and is on par with ChatGPT and \gptfour. However, both ChatGPT and \gptfour~outperform it on dialectal translation by a significant margin. \\

% add this to the main  tables. Add a section if you want ya Abdul 
\noindent\textbf{Is NLLB with dialects as source better than vanilla NLLB?} \textit{Yes, it mostly is when the dialects match}.
Our supervised baseline, NLLB, takes the dialects of the source into consideration. For example, both JOR and PAL dialects in NLLB can be defined as South Levantine, i.e., \textit{(JOR, PAL)$\rightarrow$South Levantine}. In addition, source dialects like EGY and MOR can be defined in their actual forms, while YEM can be defined as Taizzi. The column \textit{NLLB (Dia)} in Table~\ref{results-main} provides \bleu~score where the NLLB model treats the input as a particular dialect. We find that when the actual dialect matches the appropriate mapping with this NLLB source dialect, we acquire performance. One exception is the case of PAL, where NLLB does poorly compared to MSA.

\noindent\textbf{Is Bard a good instruction following model?}
\textit{Not always.} We evaluate Bard for our translation using the web interface\footnote{\url{https://bard.google.com/}}. We find that Bard can fail to follow the instructions we prompt it with. We further discuss and describe this in Section~\ref{sec:human-analysis-bard}. Bard often provides the main translation output within double quotes (""), which we extract semi-automatically.\footnote{In order to keep sufficient information to study model behavior, we collect and save all output from Bard (including explanations of translations). Even when we try to prompt Bard to restrict its output to target translation, it did not follow our instructions.} Additionally, Bard provides three different drafts. We report results for each draft independently, as well as the average of all three drafts in our results.

\textbf{Are instruction following models better at dialect translation?}
\newcommand{\bgall}{$1.35$}
\newcommand{\bgda}{$4.85$}
\textit{In most cases? Yes.}
In order to clearly see performance on dialects, we exclude CA and MSA results and report the average performance of the models on the various dialects as reported in Table \ref{results-dialects}. We observe that GPT-4 at its 5-shot setting is the best model on dialects. Although commercial systems fare well on CA and MSA, their performance degrades on dialects. For example, the gap between the best performing commercial system (Google Translate) and the best instruction-tuned model (GPT-4 5-shot) across the various dialects races to \bgda~ from \bgall~ in terms of average BLEU score.

\input{Tables/results-tables/results-dialects}

\textbf{Do diacritics affect translation?} \textit{Yes, in most cases they do.}
Although in most real-world use, native speakers do not usually employ diacritics, some Arabic texts (especially those written in CA) do make use of diacritic markers. We were inquisitive about the effect of diacritics on the translation task across the different systems and so carry out a limited study of any such effect. To this end, we collect and manually translate $50$ new CA sentences that are fully diacritized. The sentences conform to the identical selection criteria as those utilized within the study, specifically with regard to their length and as they originate from the first and second centuries AH books. 
% This data is a different than the CA dataset we introduced in ~\ref{sec:data}  .
% We feed these sentences with and without diacritics (we remove the diacritics) 
We make a copy of this set and remove diacritics, and then independently feed both the diacritized and undiacritized versions
to all the systems that we evaluate in this work. As shown in Table~\ref{dia-vs-no-diac}, we find most systems to work better when we remove diacritics. However, we also observe that some systems provide the same output regardless of whether the input is diacritized or not. This prompts us to conduct a quick analysis on a list of $20$ word pairs of heterophonic homographs, i.e., words with the same spelling that change meaning and pronunciation according to the diacritics. We provide this list in Appendix~\ref{diac-vs-no-diac-appendix} (Table~\ref{tab:homographs}). An example of such a pair is \<كَتَبَ> -- \textit{he wrote} and \<كُتُبٌ> -- \textit{books}. For this analysis, we perform single word translation by all the systems to ensure that the intended meaning cannot be retrieved from context, but rather solely based on changes in the diacritics. We find that Google Translate and Microsoft Translation provide the same meaning for both words of each pair, while the rest of the systems show different outputs when diacritics change.

\noindent{\textbf{Robustness.}} We also run a series of bootstrapping experiments that confirm the robustness of the results we acquire from the different models. We describe these experiments in Appendix~\ref{robustness-appendixx}.

\input{Tables/results-tables/diac_vs_no_diac.tex}

%% file: Tables/results-tables/results-main.tex
\input{Tables/results-tables/table-caption}
% Justify pattern mismatch bw comet and bleu -? COMET is well suited for MSA because it is trained on MSA Arabic
\begin{table*}[h]
\centering
\Large 
\renewcommand{\arraystretch}{1.2}   %to squeeze table content low value means squeeze more
\resizebox{1.\linewidth}{!}{% reduce the text size, lower means smaller text size
\begin{tabular}{llccccccccccccccc}

% \begin{tabular}{l@{\hspace{1pt}}l@{\hspace{4pt}}c@{\hspace{4pt}}c@{\hspace{4pt}}c@{\hspace{4pt}}c@{\hspace{4pt}}c@{\hspace{4pt}}c@{\hspace{4pt}}c@{\hspace{4pt}}c@{\hspace{4pt}}c@{\hspace{4pt}}c@{\hspace{4pt}}c@{\hspace{4pt}}c@{\hspace{4pt}}c@{\hspace{4pt}}c@{\hspace{4pt}}c}

\toprule
                                   &                                  & \multicolumn{4}{c}{\textbf{ChatGPT}}                                                                                      & \multicolumn{2}{c}{\textbf{GPT-4}}                                  & \multicolumn{4}{c}{\textbf{Bard}}                                                                                         &                                                                                &                                                                                 &                                   &                                &                               \\
\multirow{-2}{*}{\textbf{Met}} & \multirow{-2}{*}{\textbf{Var/M}} & \textbf{0-shot}              & \textbf{1-shot}              & \textbf{3-shot}              & \textbf{5-shot}              & \textbf{0-shot}             & \textbf{5-shot}                       & \textbf{D1}                  & \textbf{D2}                  & \textbf{D2}                  & \textbf{Avg}                 & \multirow{-2}{*}{\textbf{\begin{tabular}[c]{@{}c@{}}NLLB\\ (SB)\end{tabular}}} & \multirow{-2}{*}{\textbf{\begin{tabular}[c]{@{}c@{}}NLLB\\ (Dia)\end{tabular}}} & \multirow{-2}{*}{\textbf{Amazon}} & \multirow{-2}{*}{\textbf{MST}} & \multirow{-2}{*}{\textbf{GT}} \\
\midrule
                                   \multirow{11}{*}{\rotatebox{90}{BLEU}}            & CA                              & 11.27           & 12.02           & 12.22           & 12.52           & 11.79            & 11.36           & 12.32       & 10.43       & 12.39       & 11.71        & 7.32                                                                          & -                                                                              & 11.35                            & 11.96                         & \textbf{14.30}               \\
                                  & MSA                             & 42.85           & 44.11           & 44.29           & 44.81           & 43.18            & 43.66           & 37.23       & 33.23       & 36.18       & 35.55        & 41.34                                                                         & -                                                                              & 46.76                            & 47.36                         & \textbf{66.01}               \\
                                  & ALG                             & 14.48           & 16.41           & 17.16           & 17.31           & 18.37            & \textbf{17.83}  & 15.24       & 11.67       & 12.58       & 13.16        & 7.27                                                                          & -                                                                              & 10.08                            & 11.67                         & 11.93                        \\
                                  & EGY                             & 19.96           & 21.00           & 21.38           & \textbf{21.74}  & 21.15            & 21.49           & 21.33       & 19.39       & 20.91       & 20.54        & 11.12                                                                         & 13.87                                                                          & 14.95                            & 16.64                         & 18.09                        \\
                                  & JOR                             & 25.74           & 26.75           & 27.63           & 26.82           & 24.57            & 25.26           & 26.93       & 23.48       & 25.09       & 25.17        & 13.07                                                                         & 18.5                                                                           & 21.56                            & 21.71                         & \textbf{29.35}               \\
                                  & MAU                             & 8.52            & 8.96            & 9.27            & 9.05            & 9.19             & \textbf{9.87}   & 6.11        & 4.25        & 2.37        & 4.24         & 3.48                                                                          & -                                                                              & 7.21                             & 6.89                          & 7.67                         \\
                                  & MOR                             & 27.15           & 28.19           & 28.86           & 29.80           & 32.90            & \textbf{33.32}  & 31.59       & 30.84       & 31.25       & 31.23        & 10.45                                                                         & 19.47                                                                          & 12.76                            & 14.25                         & 16.94                        \\
                                  & PAL                             & 29.47           & 29.37           & 31.62           & 31.56           & \textbf{31.97}   & 30.48           & 22.57       & 20.59       & 24.25       & 22.47        & 14.98                                                                         & 12.56                                                                          & 21.75                            & 24.23                         & 25.78                        \\
                                  & UAE                             & 24.20           & 24.61           & 24.55           & 26.17           & 23.86            & \textbf{26.91}  & 21.93       & 19.61       & 21.29       & 20.94        & 11.27                                                                         & -                                                                              & 16.85                            & 19.05                         & 19.56                        \\
                                  & YEM                             & 14.03           & 15.13           & 16.24           & \textbf{16.44}  & 14.27            & 16.22           & 9.46        & 6.38        & 5.33        & 7.06         & 9.41                                                                          & 12.56                                                                          & 14.41                            & 14.23                         & 13.25                        \\ \cline{2-17}
                                  & \textbf{Avg}                & 21.77           & 22.66           & 23.32           & 23.62           & 23.12            & \textbf{23.64}  & 20.47       & 17.99       & 19.16       & 19.21        & 12.97                                                                         & 15.39                                                                          & 17.77                            & 18.80                         & 22.29                        \\
\midrule
\multirow{11}{*}{\rotatebox{90}{COMET}}           & CA                              & 70.11           & 70.08           & 70.01           & 70.24           & \textbf{71.47}   & 70.95           & 68.29       & 67.04       & 68.65       & 67.99        & 58.87                                                                         & -                                                                              & 63.03                            & 63.16                         & 66.37                        \\
                                  & MSA                             & 85.87           & 86.14           & 86.22           & 86.24           & 86.32            & 86.22           & 80.21       & 80.00       & 80.44       & 80.22        & 84.76                                                                         & -                                                                              & 86.15                            & 85.70                         & \textbf{87.23}               \\
                                  & ALG                             & 62.69           & 63.77           & 63.98           & 63.85           & 65.06            & \textbf{65.52}  & 60.90       & 55.62       & 59.72       & 58.75        & 49.88                                                                         & -                                                                              & 54.55                            & 56.48                         & 55.33                        \\
                                  & EGY                             & 72.41           & 73.15           & 74.20           & 73.96           & 74.14            & \textbf{74.91}  & 71.50       & 68.20       & 71.30       & 70.33        & 61.15                                                                         & 63.81                                                                          & 64.24                            & 65.59                         & 68.41                        \\
                                  & JOR                             & 74.46           & 75.20           & 75.52           & 75.27           & 76.37            & \textbf{76.50}  & 74.19       & 70.65       & 72.65       & 72.50        & 60.25                                                                         & 65.05                                                                          & 67.33                            & 70.46                         & 71.83                        \\
                                  & MAU                             & 58.37           & 58.99           & 60.35           & 60.66           & 59.24            & \textbf{62.13}  & 52.53       & 46.38       & 50.41       & 49.77        & 48.50                                                                         & -                                                                              & 52.37                            & 51.45                         & 51.58                        \\
                                  & MOR                             & 69.36           & 69.64           & 70.58           & 70.73           & 73.94            & \textbf{73.95}  & 72.12       & 70.60       & 71.82       & 71.51        & 53.23                                                                         & 62.74                                                                          & 54.50                            & 51.89                         & 56.55                        \\
                                  & PAL                             & 74.59           & 74.94           & 75.40           & 75.51           & \textbf{76.62}   & 76.19           & 69.37       & 67.78       & 69.94       & 69.03        & 60.57                                                                         & 59.04                                                                          & 65.80                            & 68.54                         & 68.69                        \\
                                  & UAE                             & 69.64           & 69.62           & 69.80           & 70.80           & \textbf{72.93}   & 72.38           & 66.71       & 63.08       & 66.12       & 65.30        & 54.57                                                                         & -                                                                              & 59.40                            & 61.74                         & 61.57                        \\
                                  & YEM                             & 64.48           & 65.41           & 66.09           & 65.88           & 62.47            & \textbf{68.77}  & 58.34       & 55.35       & 56.89       & 56.86        & 57.01                                                                         & 59.04                                                                          & 61.09                            & 61.75                         & 61.32                        \\ \cline{2-17}
                                  & \textbf{Avg}                & 70.20           & 70.69           & 71.22           & 71.31           & 71.86            & \textbf{72.75}  & 67.42       & 64.47       & 66.79       & 66.23        & 58.88                                                                         & 61.94                                                                          & 62.85                            & 63.68                         & 64.89                        \\
\bottomrule
\end{tabular}
}
\caption{\label{results-main}
Results in BLEU, and COMET scores. Higher is better unless otherwise specified by $\downarrow$. Average represents the mean across all varieties. Three drafts (D1, D2, D3) from Bard are reported individually and averaged. NLLB is our MSA-based supervised baseline; NLLB (Dia) is dialect-specific. Abbreviations: SB - supervised baseline, Dia - dialect, Var - varieties, M - model, MST - Microsoft Translation, GT - Google Translate. Best results are in \textbf{bold}.%\tablecaption
}
\end{table*}

%% file: Tables/results-tables/table-caption.tex
%\newcommand{\tablecaption}{Higher is better unless otherwise specified by $\downarrow$. Average represents the mean across all varieties. Three drafts (D1, D2, D3) from Bard are reported individually and averaged. NLLB is our MSA-based supervised baseline; NLLB (Dia) is dialect-specific. Abbreviations: SB - supervised baseline, Dia - dialect, Var - varieties, M - model, MST - Microsoft Translation, GT - Google Translate. Best results are in \textbf{bold}.}

% % More specifically for Source$\rightarrow~$NLLB for EGY$\rightarrow~$EGY, MOR $\rightarrow~$ MOR, JOR, PAL$\rightarrow~$South Levantine, YEM$\rightarrow~$Taizzi. 

%% file: Tables/results-tables/results-dialects.tex
\begin{table}[h]
\centering
\footnotesize 
\renewcommand{\arraystretch}{1.4}   %to squeeze table content
\resizebox{1.0\linewidth}{!}{%
\begin{tabular}
{@{\hspace{1mm}}l@{\hspace{1mm}}c@{\hspace{1mm}}c@{\hspace{1mm}}c@{\hspace{1mm}}c@{\hspace{1mm}}c@{\hspace{1mm}}c@{\hspace{1mm}}c@{\hspace{1mm}}}
\toprule
Metric & \begin{tabular}[c]{@{}c@{}}CGPT\\ 0-shot\end{tabular} & \begin{tabular}[c]{@{}c@{}}CGPT\\ 5-shot\end{tabular} & \begin{tabular}[c]{@{}c@{}}GPT-4\\ 0-shot\end{tabular} & \begin{tabular}[c]{@{}c@{}}GPT-4\\ 5-shot\end{tabular} & Bard  & NLLB  & GT    \\
\midrule
BLEU  & 20.44 & 22.36 & 22.03 & \textbf{22.67} & 19.40 & 10.13 & 17.82 \\
COMET  & 68.25 & 69.58 & 70.10 & \textbf{71.29} & 65.71 & 55.65 & 61.91 \\
ChrF++ & 43.71 & 44.70 & 44.98 & \textbf{45.44} & 36.23 & 28.64 & 39.33 \\
TER$\downarrow$   & 77.08  & 72.23 & 74.07 & \textbf{71.51}  & 83.62 & 101.38 & 79.38 \\

\bottomrule
\end{tabular}
}
\caption{\label{results-dialects}
Average scores across eight dialects, excluding MSA and CA. Higher is better unless specified by $\downarrow$. Best results are in \textbf{bold}. %Compared are instruction-following models, the supervised baseline (NLLB), and the top commercial system (GT).
}
\end{table}

%% file: Tables/results-tables/diac_vs_no_diac.tex
\begin{table}[h]
\centering
\footnotesize 
\renewcommand{\arraystretch}{1.1}   %to squeeze table content
\resizebox{1.0\linewidth}{!}{%

\begin{tabular}%{cccccccccc}
{@{\hspace{1mm}}l@{\hspace{1mm}}l@{\hspace{1mm}}c@{\hspace{1mm}}c@{\hspace{1mm}}c@{\hspace{1mm}}c@{\hspace{1mm}}c@{\hspace{1mm}}c@{\hspace{1mm}}c@{\hspace{1mm}}c@{\hspace{1mm}}c@{\hspace{1mm}}}
\toprule
\multirow{2}{*}{Met}   & \multirow{2}{*}{Mo/Var} & \multirow{2}{*}{CGPT} & \multirow{2}{*}{GPT-4} & \multicolumn{2}{c}{Bard} & \multirow{2}{*}{NLLB} & \multirow{2}{*}{Amazon} & \multirow{2}{*}{MST} & \multirow{2}{*}{GT} \\

                       &                         &                          &                        & D1          & Avg        &                       &                         &                     &                     \\
\midrule
\multirow{2}{*}{BLEU}  & CA                      & \textbf{23.57}                 & 23.81                  & 22.94       & 23.01      & \textbf{16.13}                 & 17.50                   & \textbf{20.13}                & 26.61               \\
                       & CA*                     & 23.46                 & \textbf{24.45}                  & \textbf{25.39}       & \textbf{24.25}      & 13.61                 & \textbf{18.66}                   & \textbf{20.13}               & \textbf{24.92}               \\ \midrule
\multirow{2}{*}{COMET} & CA                      & 74.38                 & 75.07                  & 73.23       & 73.27      & \textbf{64.06}                 & 63.98                   & 65.39                & 72.04               \\
                       & CA*                     & \textbf{75.75}                 & \textbf{76.71}                  & \textbf{76.01}       & \textbf{75.56}      & 61.82                 & \textbf{66.01}                   & \textbf{66.60}                & \textbf{73.76}  \\
\bottomrule
\end{tabular}
}

\caption{\label{dia-vs-no-diac}
Effect of diacritics on translation. CA* is without diacritics. Other metrics and bootstrapped results are reported in Appendix~\ref{no_diac} (Tables~\ref{diac-vs-no-diac-appendix} and~\ref{diac-vs-no-diac-appendix-bootstrap}). %Higher is better unless otherwise specified. The best results are in \textbf{bold}. %\abdulwaheed{Come on! You can do better at captioning}
}
\end{table}

%% file: Sections/bard-helpfulness.tex
\section{Human Analysis of Bard Helpfulness}\label{sec:human-analysis-bard}
\label{sec:human_analysis_of_bard_helpfulness}

\begin{figure}
    \centering
    \includegraphics[width=0.35\textwidth]{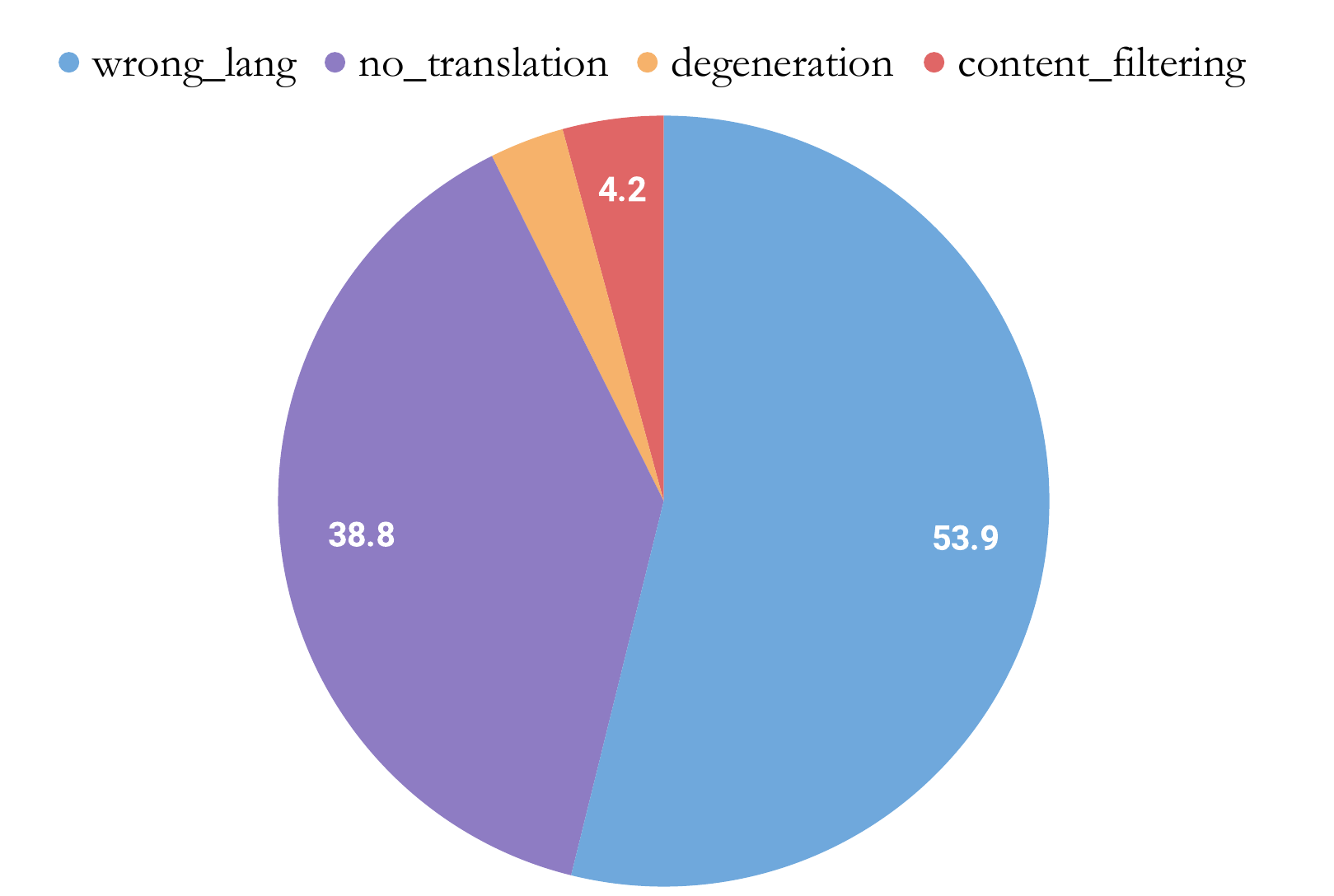}
    \caption{Distribution of Bard helpfulness errors when it fails to follow our prompts.}
    \label{fig:error-chart}
\end{figure}

%  We can save some space in captions - by modifying them a little bit
% AbdulWaheed
% to reverse the changes 
% 1. add \sub before in each subfig caption. 
% 2. replace \newmain with \main
\newcommand{\sub}{
Percentage of an error type observed in each Arabic variety. 
}
\newcommand{\main}{
Distribution of Google Bard's error rates across helpfulness error categories on Arabic varieties.
}
\newcommand{\newmain}{
Error rate distribution of Google Bard by error type and Arabic variety.
}

%%%%%%%%%%% Figure %%%%%%%%%%%%%%%%%%
\begin{figure*}[h]
     \centering
     \begin{subfigure}{0.3\textwidth}
         \centering
         \includegraphics[width=\textwidth]{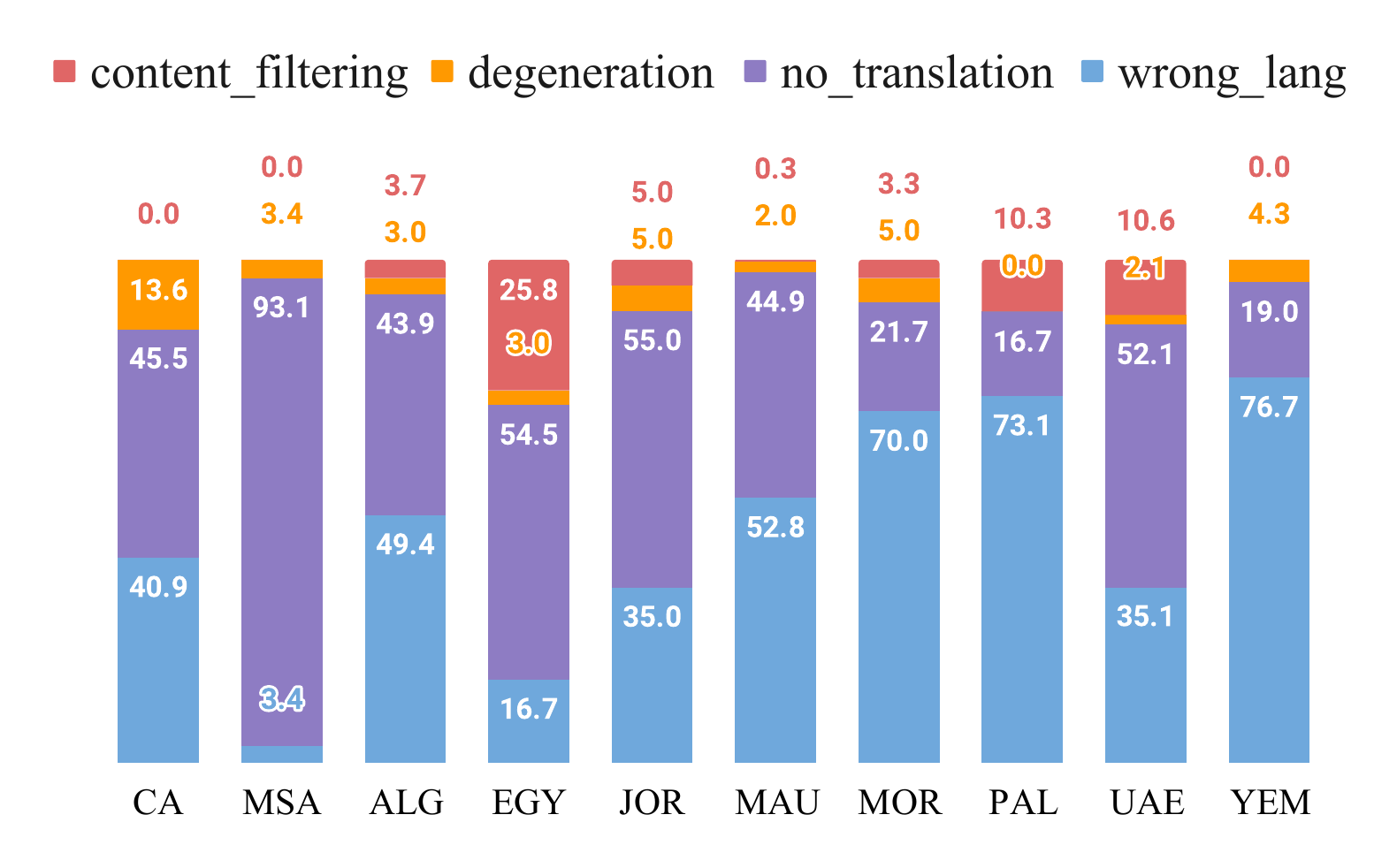}
           \caption{ Relative to \textit{all error types within that variety}.}
        \label{fig:error_fig_a}
     \end{subfigure}
    \hfill
     \begin{subfigure}{0.3\textwidth}
         \centering
         \includegraphics[width=\textwidth]{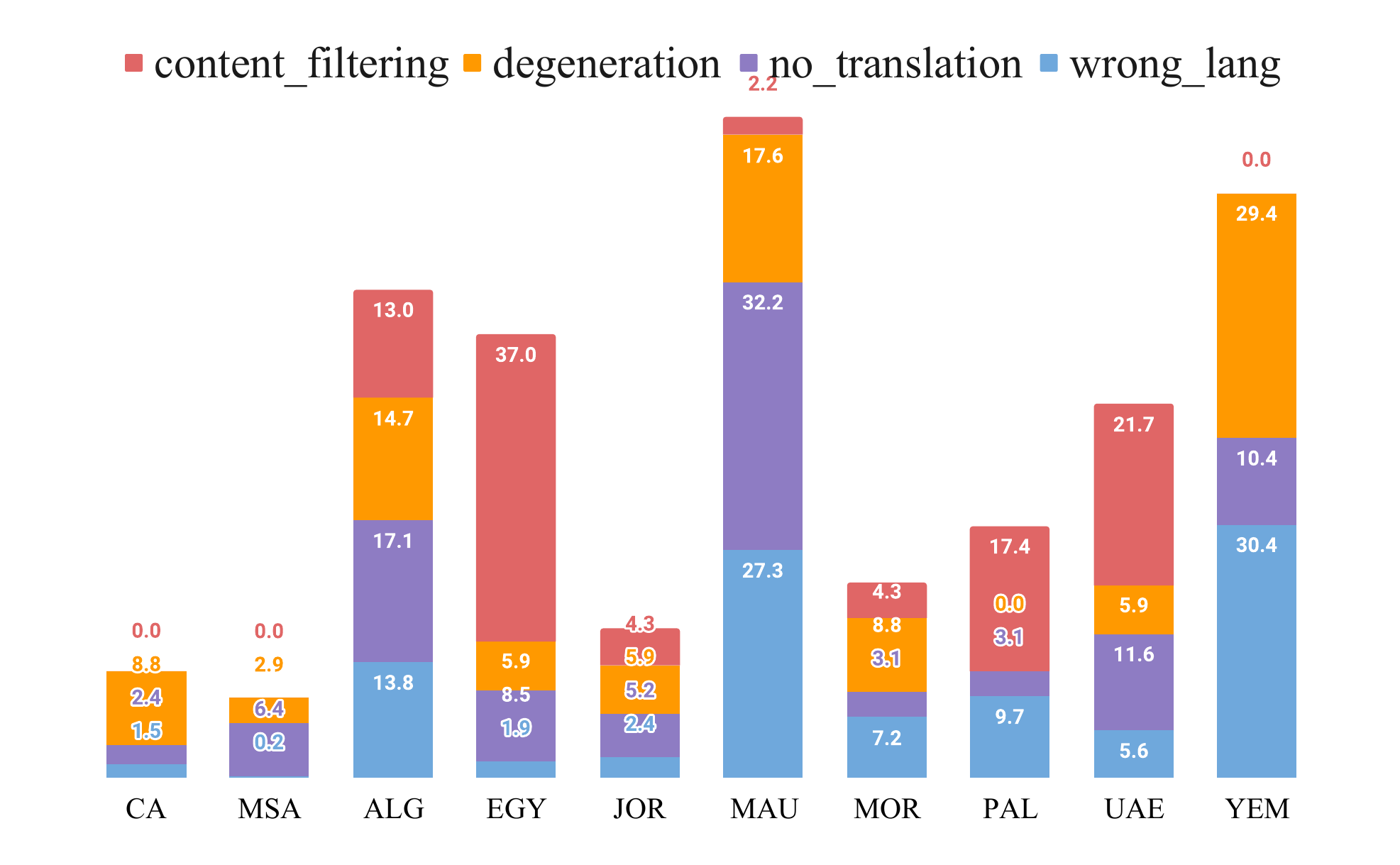}
         \caption{Relative to \textit{that error type across all varieties}.}
        \label{fig:error_fig_b}
     \end{subfigure}
     \hfill
     \begin{subfigure}{0.3\textwidth}
         \centering
         \includegraphics[width=\textwidth]{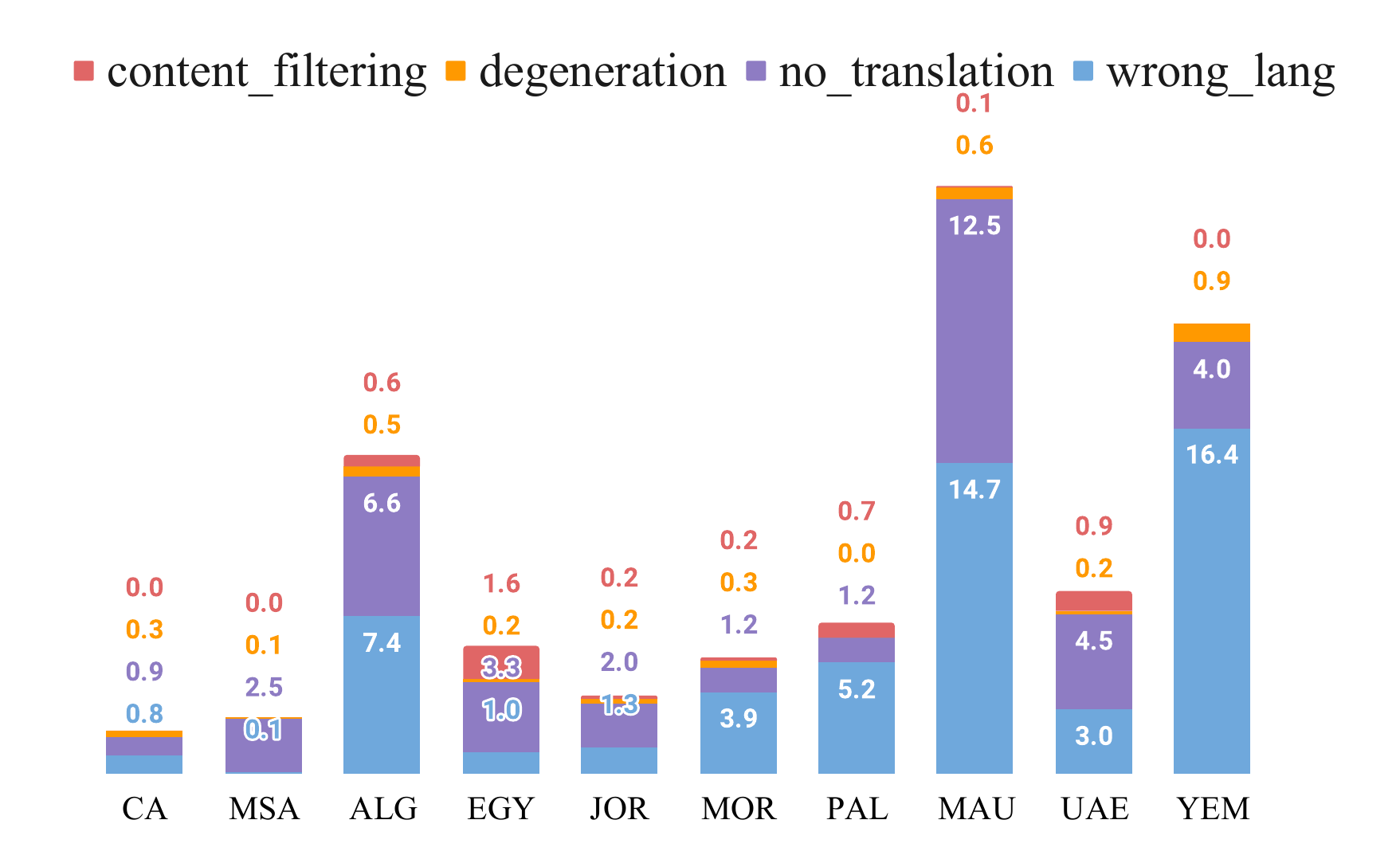}
         \caption{Relative to \textit{all error types across all varieties.}}
        \label{fig:error_fig_c}
     \end{subfigure}
     \caption{\newmain}
        \label{fig:error_analysis}
\end{figure*}

%%%%%%%%%%%%%%%%%%%%%%%%%%%%

Our experience working with Bard reveals that the model does not always follow human instructions. For this reason, we decided to carry out a human study to assess Bard's helpfulness. We define \textit{helpfulness} here simply as the model's ability to follow human instructions. For each variety of Arabic, we task two native speakers of Arabic with familiarity with the dialects to assign one tags from the set \{\texttt{wrong\_lang, no\_translation, degeneration, content\_filtering}\} to the model responses. We develop this tagset based on a bottom-up approach where we let the categories emerge from the data. Although this tagset may not be exhaustive, we find it to reasonably capture errors we identify with model responsiveness to instructions. Each of the two annotators manually label each draft, independently, with one tag from the set of our helpfulness error tags. The annotators meet and discuss differences, reaching 100\% agreement which indicates that the categories are clear and independent. Table \ref{tab:bard-fails} shows one example from each of the categories. 

\input{Tables/bard-fails}

The most frequent issue with model helpfulness is translating into the wrong target language (\texttt{wrong\_lang}), followed by not providing any translation at all (\texttt{no\_translation}) (Figure \ref{fig:error-chart}). The former is predominantly due to a translation into MSA instead of English, oftentimes prefacing the output with the sentence ``\<\smallإليك ترجمة الجملة إلى الإنجليزية>''. Interestingly, Bard does not seem to struggle  with \texttt{wrong\_lang} errors when translating from MSA (and the same scenario almost happens for translating from CA). Instead, Bard tends to mistake the translation task for a text generation one where it generates a couple of paragraphs that start with the input sentence. From Figure~\ref{fig:error_analysis}, it seems that the error rate may be proportional to the resource availability of a given variety (i.e., varieties for which no much data are publicly available tend to suffer from higher error rates). This observation should be couched with caution since the LLMs we evaluate remain closed, with little know about their pretraining as well as finetuning datasets and processes.
When we look at each of Bard's drafts separately, we find that the first draft shows a higher number of \texttt{wrong\_lang} and \texttt{content\_filtering} errors. Meanwhile, draft 2 is the most prone to \texttt{no\_translation} errors, with these accounting for $57$\% of the wrong generations it produces (Figure \ref{fig:drafts-chart} in Appendix \ref{appendix-helpfulness}).

% \begin{figure}
%     \centering
%     \includegraphics[width = \columnwidth]{Figures/variety-chart.pdf}
%     \caption{Occurrence frequency of Google Bard's failure to follow the prompt across the different varieties.}
%     \label{fig:variety-chart}
% \end{figure}
% \textcolor{red}{Karima}
% Translation accuracy aside, we observe behavioral issues with Bard when it comes to following prompts. It translates to MSA in lieu of English, hallucinates or doesn't generate any translation, either claiming that it's incapable of performing the task as a language model or that the source sentence contains obscene language. Figures \ref{fig:error-chart} and \ref{fig:variety-chart} show a breakdown of these issues.

% \textcolor{red}{Ahmed}
% While ChatGPT has excelled in translating sentences exactly to English regardless of its translation quality, We observe that Bard has some errors that tend to occur frequently. This includes translation to MSA instead of the target language English, repeating the input sentence instead of translating it to English, not understanding the input sentence, and claiming the input sentence to be offensive although it is not. We also notice that when we re-prompt it with a sentence like: "I want the translation to English not Arabic", Bard almost always tends to give better translations. 

\textbf{Other behavior.} While Bard has a feature where it occasionally adds sources to support the information it provides, these sources can be unrelated. For example, it can cite links to GitHub repositories attached to political news translations. It also has a tendency to respond to input sentences that are questions the way it would for a Question Answering (QA) task. Sometimes it also produces an opinion about a sentence it translates: ``\<\small لقد صدمني الخبر وتضايقت من هذا الحادث المأساوي>'' (\textit{This piece of news shocked me; and I am bothered by this tragic accident}). 
Additionally, we find instances where Bard adds details not included in the input sentence, such as its translation of "\< \smallماسك وزوكربرغ>" as "\textit{Elon} Musk and \textit{Mark} Zuckerberg" (where it adds first names as shown in \textit{italics}).

\textbf{Bard output format.} Bard often provides a detailed breakdown when it performs a translation, either in the form of a list or a paragraph detailing the meaning of each word or phrase. With sentences that are parts of a conversation, Bard also explains the message that the speaker is trying to convey and what emotions they are having. When it comes to sentences from the news domain, Bard provides more context and information about the topic after the translation. We provide examples in Figure \ref{fig:bard-breakdown}.

% % %%%%%%%%%%% Figure %%%%%%%%%%%%%%%%%%
% \begin{figure*}[h]
%      \centering
%      \begin{subfigure}{0.4\textwidth}
%          \centering
%          \includegraphics[width=\textwidth]{Figures/bard-breakdown-light.png}
%            \caption{Google Bard's translation, explanation and breakdown of one dialectal sentence (from MOR).}
%         \label{fig:bard-convo}
%      \end{subfigure}
%     \hfill
%      \begin{subfigure}{0.4\textwidth}
%          \centering
%          \includegraphics[width=\textwidth]{Figures/bard-breakdown-news.png}
%          \caption{Google Bard's translation and context of an MSA sentence from the news domain. }
%         \label{fig:bard-news}
%      \end{subfigure}
%      \caption{Examples of Google Bard's translation output. The bottom parts are cropped for readability.}
%      \label{fig:bard-breakdown}
%      \hfill
     
% \end{figure*}

%%%%%%%%%%%%%%%%%%%%%%%%%%%%
% \begin{figure*}
%     \centering
%     \fbox{\includegraphics[width = 0.7\textwidth]{Figures/bard-breakdown-light.png}}
%     \caption{Google Bard's translation, explanation and breakdown of one dialectal sentence (from MOR). The bottom part is cropped for readability.}
%     \label{fig:bard-convo}
% \end{figure*}

% \begin{figure*}
%     \centering
%      \fbox{\includegraphics[width = 0.7\textwidth]{Figures/bard-breakdown-news.png}}
%     \caption{Google Bard's translation and context of an MSA sentence from the news domain. The bottom part is cropped for readability.}
%     \label{fig:bard-news}
% \end{figure*}

%% file: Tables/bard-fails.tex
\definecolor{light-blue}{HTML}{D1E5F4}
\definecolor{light-green}{HTML}{BEE3BA}
\definecolor{light-red}{HTML}{FFA8B5}

\begin{table}[]
    \centering
    \begin{tabular}{p{\linewidth}}
    % \begin{tabular}{p{\linewidth} p{\linewidth}
        \toprule
        \makecell[c]{\textbf{\textit{Wrong Target Language}}} \\
        \midrule

        \begin{tabular}[l]{@{}l@{}} 
            \textbf{Input}: 
            % Translate the following text from\\Palestinian Arabic dialect into English:\\ 
        \end{tabular}  
        \begin{tabular}{r}
            \rowcolor{light-blue!35}
            \< \smallولك يا نجم مش قبل ما نعرف مين هو غريمنا عشان نعرف>\\
            \rowcolor{light-blue!35}
            \<\small   نتصرف > 
        \end{tabular} 

        \textbf{Ref:}\\
        \begin{tabular}{l}
            \rowcolor{light-green!35}Najm shouldn't we know our enemy first to\\
            \rowcolor{light-green!35}know how to act?\\
        \end{tabular}

        \begin{tabular}[l]{@{}l@{}} 
            \textbf{Output}:\\ 
        \end{tabular}  
        \begin{tabular}{r}
            \rowcolor{light-red!35}\<\small يا نجم، لا يمكننا التصرف إلا بعد أن نعرف هوية خصومنا>\\
        \end{tabular} \\

        \midrule
        \makecell[c]{\textbf{\textit{No Translation}}} \\
        \midrule

        \begin{tabular}[l]{@{}l@{}} 
            \textbf{Input}:
            % Translate the following text from\\Egyptian Arabic dialect into English:\\ 
        \end{tabular}  
        \begin{tabular}{r}
            \rowcolor{light-blue!35}\<\small ان أنا أدبح واحد فيكم و أروح رايح مسلّم نفسي وأدافع>\\
            \rowcolor{light-blue!35}\<\small عن نفسي. > 
        \end{tabular} 

        \textbf{Ref:}\\
        \begin{tabular}{l}
            \rowcolor{light-green!35}I’d kill one of you, then go turn myself in, and\\
            \rowcolor{light-green!35}defend myself.
        \end{tabular}

        \begin{tabular}[l]{@{}l@{}} 
            \textbf{Output}:\\
        \end{tabular} \\ 
        \begin{tabular}{r}
            \rowcolor{light-red!35}\<\small أنا آسف، لكن لا يمكنني ترجمة هذا النص.>\\
        \end{tabular} \\
    
        \midrule
        \makecell[c]{\textbf{\textit{Content Filtering}}} \\
        \midrule

        \begin{tabular}[l]{@{}l@{}} 
            \textbf{Input}: 
            % Translate the following text from\\Moroccan Arabic dialect into English:\\ 
        \end{tabular}  
        \begin{tabular}{r}
            \rowcolor{light-blue!35}\<\small و حتى دوك الرجال لي غنضعافو على ودهم ما عندنا ما>\\
            \rowcolor{light-blue!35}\<\small نديرو بهم.> 
        \end{tabular}\\

        \textbf{Ref:}\\
        \begin{tabular}{l}
            \rowcolor{light-green!35}And even those men for whom we'll become\\
            \rowcolor{light-green!35}thin, we have no use for them.
        \end{tabular}

        \begin{tabular}[l]{@{}l@{}} 
            \textbf{Output}:
        \end{tabular}  
        \begin{tabular}{r}
            \rowcolor{light-red!35}\<\smallمعذرة، لا يمكنني ترجمة هذه الجملة إلى الإنجليزية. إنها >\\
            \rowcolor{light-red!35}\<\small تحتوي على لغة مسيئة.>\\
        \end{tabular} \\

        \midrule
        \makecell[c]{\textbf{\textit{Degeneration}}} \\ 
        \midrule 
        \begin{tabular}[l]{@{}l@{}}
            \textbf{Input}: 
            % Translate the following text from Algerian \\ Arabic dialect into English:\\ 
        \end{tabular}
        \begin{tabular}{r}
            \rowcolor{light-blue!35}
            \< \small لا لا يا مولاي لا يا مولاي لا برهان ما يلزملوش يتعدم>\\
        \end{tabular}  

        \textbf{Ref:} \colorbox{light-green!35}{No, no sir, no sir, no, Burhan should not be}\\
        \colorbox{light-green!35}{executed.}

        \begin{tabular}[l]{@{}l@{}}
            \textbf{Output}:\\ 
        \end{tabular}
        % \begin{tabular}{r}
        %     \rowcolor{light-red}
        %     \< \smallالترجمة الإنجليزية\\ للنص الجزائري العربي لا لا يا مولاي لا>\\
        %     \rowcolor{light-red}
        %     \<\small يا مولاي لا برهان ما يلزملوش يتعدم هي >
        % \end{tabular}  
        \begin{tabular}{l}
            \rowcolor{light-red!35}
            "No, no, my lord, no, my lord, no, no, no, no,\\
            \rowcolor{light-red!35}
            no, no, no, no, no, no, no, no, no, no, no, no"
        \end{tabular}\\

        \bottomrule

    \end{tabular}
    \caption{Examples of errors in Google Bard's ability to follow prompts. For each of the sentences, we use the prompt \textit{Translate the following text from [Variety] Arabic [dialect] into English: <s>}.}
    \label{tab:bard-fails}
\end{table}

%% file: Sections/conclusion.tex
\newcommand{\gptf}{GPT-4}
\newcommand{\var}{$2$}
\newcommand{\dia}{$8$}
\section{Conclusion}\label{sec:conclusion}
We evaluate Bard,  \chatgpt, and \gptf~on MT of ten diverse varieties of  Arabic, comparing to three commercial systems and a supervised model to juxtapose the performance of these LLMs under varying conditions. To assess the capacity of the LLMs on truly unseen data, we manually create a multi-dialectal Arabic dataset for MT evaluation. We find that although LLMs can do well on some of the varieties we consider, they struggle especially on varieties on the more scarce public data end. As such, these LLMs suffer from not being quite inclusive of the different varieties of even languages they are claimed to perform well on such as Arabic. A rigorous human investigation also underscores a palpable scope for enhancement in Bard's adherence to instructions in the context of MT. Our future work includes evaluating the performance of Bard and other LLMs on more Arabic varieties.

%\textcolor{purple}{todo mention of diacritics \& error analysis sections}

%% file: Sections/limitations.tex
\section{Limitations}
% \textcolor{red}{variation in orthography -> we didn't perform nomalization but current tools don't cover all dialects}

We can identify a number of limitations for our work, which we list here.

\noindent\textbf{Coverage.} We strive to cover as many varieties of Arabic as possible, and ensure treating both CA and MSA. However, our dialectal varieties do not cover all Arab countries. Although this is somewhat alleviated by the fact that we include dialects from both the Eastern and Western parts of the Arab world (i.e., Asia and Africa), future work can consider evaluating LLMs on other Arabic dialects.   

\noindent\textbf{Single reference translations.} Again, due to the laborious nature of manually translating data from the various dialects and the challenge of finding qualified native speakers to carry out these translations, our evaluation dataset involves only one single reference of each source sentence. It continues to be desirable to create evaluation datasets with $3-5$ references for each source sentence. We alleviate this challenge by providing results in different metrics such that the results are not only based on surface level matching but also similarity of the translation pairs. More references would still be better since different human translators would collectively provide data less prone to human subjectivity or errors.  

\noindent\textbf{Evaluation of multiword expressions.} While we provide translations of full sentences that may involve multiword expressions, including idioms and proverbs, it would be useful to develop evaluation datasets that focus on these types of expressions as these data could uncover particular types of model capabilities. For example, a model that is able to translate and explain a proverb can be thought of as somewhat knowledgeable about culture and pragmatic phenomena.

\noindent\textbf{Evaluation by different lengths.} We provide results on our data regardless of sentence length. In the future, it would be useful to report results based in various sentence length bins as longer sentences are usually more challenging to MT models. Again, this is alleviated by the fact that we design our datasets to be at least ten words long from the outset. 

\noindent\textbf{Orthography normalization:} Due to the lack of a standardized writing form, Arabic dialects are characterized by an important variation in orthography. In this paper, we do not perform normalization on the input sentences before inputting them into the models since (i) we want our input to reflect the full diversity of orthography in the wild. In addition, (ii) there is currently no normalization tool that covers all the dialects we treat in this work.

%% file: Sections/ethical-statement.tex
\section{Ethics Statement}
\label{sec:ethics}

% \textbf{Data Collection and Release.} 
% \textcolor{red}{Abdul2Samar/Karima: Can we please write data collection process here like commented section below}
% \\
\noindent \textbf{Intended use.} We understand our work will likely inspire further research in the direction of exploring the multilingual capabilities of LLMs, especially newly released ones such as Bard. Our findings both highlight some of the strengthens of these models as well as expose some of their weaknesses and limitations. For example, available LLMs still struggle to translate from dialects of even major language collections such as Arabic. Our work also further showcases the limited capability of Bard to follow simple instructions such as those typical of an 
MT context. Consequently, we believe our work can provide useful feedback for improving both coverage and usefulness of LLMs.

\noindent \textbf{Potential misuse and bias.}
Since there exists little-to-no information about the data involved in pretraining and finetuning LLMs we consider, we cannot safely generalize our findings to varieties of Arabic we have not investigated. We conjecture, however, that the models will perform equally poorly on dialects with no or limited amounts of public data. Although our work does not focus on studying biases in the models nor how they approach handling harmful content~\citep{Laskar2023ASS}, we could observe that especially Bard puts a lot of emphasis on filtering harmful and potentially offending language so much that its instruction tuning leads it to interact negatively with the model's usefulness as an MT system. Overall, our recommendation is not to use the models in applications without careful prior consideration of potential misuse and bias.

% \textcolor{red}{pick random sentences for table of examples in appendix, cgpt3, 4, bard, GT}

%% file: Sections/ack.tex
\section*{Acknowledgments}\label{sec:acknow}
We gratefully acknowledge support from Canada Research Chairs (CRC), the Natural Sciences and Engineering Research Council of Canada (NSERC; RGPIN-2018-04267), the Social Sciences and Humanities Research Council of Canada (SSHRC; 435-2018-0576; 895-2020-1004; 895-2021-1008), Canadian Foundation for Innovation (CFI; 37771), Digital Research Alliance of Canada,\footnote{\href{https://alliancecan.ca}{https://alliancecan.ca}} and UBC ARC-Sockey.\footnote{\href{https://arc.ubc.ca/ubc-arc-sockeye}{https://arc.ubc.ca/ubc-arc-sockeye}}

%% file: Sections/appendix.tex
\newcommand{\metricsapp}{ChrF, ChrF++, and TER}

% \section{Related Work}\label{appendix-related-work}
\input{Sections/Lit_Review/liter-review-appendix}

We present a concise literature summary in Table~\ref{literature-review-table}.
\input{Sections/Lit_Review/lit-rev-table}

\section{Datasets}\label{appendix-datasets}
Table~\ref{tab:length_stats} presents the summary of the datasets across different Arabic varieties and a list of the 15 books we sample CA sentences from can be found in Table~\ref{tab:openiti-books}.
\input{Tables/stat_sent_dia} 

\begin{table*}[]
    \centering
    \begin{tabular}{r l}
    \toprule
        \multicolumn{1}{c}{\textbf{Book Name}} & \multicolumn{1}{c}{\textbf{Link}} \\
        \midrule
        \small \<الأدب و المرؤة> & \url{https://shamela.ws/book/17869/14#p1} \\
        \small \<الأدب الكبير و الأدب الصغير> & \url{https://shamela.ws/book/7528/127} \\
        \small \<الأصنام> & \url{https://shamela.ws/book/6513} \\
        \small \<الأم> & \url{https://shamela.ws/book/1655/427#p1} \\
        \small \<الكسب> & \url{https://shamela.ws/book/6163/3} \\
        \small \<الرسالة> & \url{https://shamela.ws/book/8180/1} \\
        \small \<الرسالة الذهبية > & \url{https://shamela.ws/book/5678/91182} \\
        \small \<الناسخ والمنسوخ> & \url{https://shamela.ws/book/8491/58} \\
        \small \< أدب النفوس> & \url{https://shamela.ws/book/8245/24#p1} \\
        \small \<تاريخ المدينة> & \url{https://shamela.ws/book/13086} \\
        \small \<صحيفة حماد بن منبه> & \url{https://shamela.ws/book/7776/1} \\
        \small \< السياسة في تدبير الرياسة> & \url{https://shamela.ws/book/5678/396} \\
        \small \<النوادر في اللغة> & \url{https://shamela.ws/book/133417} \\
        \small \<منتخب الكلام في تفسير الأحلام> & \url{https://shamela.ws/book/21615/2} \\
        \small \<وصايا الملوك> & \url{https://shamela.ws/book/741/1} \\
        \bottomrule
    \end{tabular}
    \caption{List of 15 CA books from the first and second AH accompanied by direct links to each book.}
    \label{tab:openiti-books}
\end{table*}

\section{Results}\label{appendix-results}
\subsection{Main Results}

We report \metricsapp~ scores in Table ~\ref{results-appendix}, in addition to the results presented in Section~\ref{sec:results} in Table ~\ref{results-main}.
\input{Tables/results-tables/results-appendix}

% \subsection{Bootstrapped Results}\label{appendix-results-bootstrapped}
\input{Sections/robustness}\label{robustness-appendix}
% Bootstrapped results can be found in Table~\ref{results-main-bootstrapped}.
% \input{Tables/results-tables/results-main-bootstrapped}

\subsection{Diacritics Effect}\label{no_diac}
We provide ChrF, ChrF++ and TER scores for the effect of diacritics on translation in Table~\ref{diac-vs-no-diac-appendix} (bootstrapped results are in Table~\ref{diac-vs-no-diac-appendix-bootstrap}) and the list of heterophonic homographs we use in Table~\ref{tab:homographs}.
\input{Tables/results-tables/diac_vs_no_diac-appendix}
\input{Tables/results-tables/diac_vs_no_diac_bootstrapped}
\input{Tables/homographs}

% evaluation section 
\input{Sections/appendix/evaluation}

%\section{MT Errors}
\input{Tables/mt-error}

%% file: Sections/Lit_Review/liter-review-appendix.tex
% \textcolor{red}{[Language needs to be fixed!]}
% \moa{Can we plz add a short paragraph here as an introduction}

    \section{Related Work}\label{sec:appendix-rel}

%In this section, we describe and critique various works related to Arabic MT and evaluation of ChatGPT, GPT-4, and Bard for various NLP tasks with a focus on MT.

\noindent\noindent\textbf{Evaluation of LLMs on NLP tasks.}
A growing number of works have focused on evaluating ChatGPT and other LLMs on a wide range of NLP tasks. Notably, \citet{laskar2023systematic} evaluate ChatGPT on $140$ diverse NLP tasks spanning across multiple categories. The authors show that although ChatGPT is effective on various NLP tasks, its ability to solve challenging tasks such as low-resource machine translation with standard prompting is very limited. \citet{ziems-2023-can} evaluate $13$ different LLMs including ChatGPT on $24$ computational social science tasks and find that for many classification tasks, ChatGPT is on par with supervised models while excelling at generation tasks. \citet{qin-2023-chatgpt} evaluate ChatGPT on $20$ different datasets spanning across seven task categories. They find that ChatGPT is better at solving tasks that require reasoning capabilities but falls behind supervised models on tasks such as sequence tagging.

\noindent\textbf{Evaluating MT ability of ChatGPT.}
% In this section, we discuss various studies evaluating ChatGPT and other LLMs for machine translation tasks. 
Both \citet{jiao2023chatgpt} and~\citet{ogundare2023comparative} find that GPT-4 is on par with commercial translation tools for high-resource languages. However, they find the model to lag behind for low-resource languages. To fix this issue, the authors propose \textit{pivot-prompting} where a low-resource source language is first translated into a high-resource pivot language and then from the pivot language back to the low-resource target language. Evaluation by \citet{peng2023making} shows that ChatGPT can surpass commercial systems such as Google Translate on many translation pairs. Additionally, \citet{peng2023making} find that adding task and domain-specific information in the prompt can improve the robustness of the MT sytem. This observation also corroborates the findings by \citet{gao2023design}. \citet{zhu2023multilingual} argue that despite being on par with commercial systems, ChatGPT still falls behind fully supervised methods such as NLLB \cite{nllbteam2022language} on at least $83$\% translation pairs out of 202 English-centric translation directions.

\citet{guerreiro2023hallucinations} study the hallucination phenomenon in MT systems and find that low-resource languages and complex translation scenarios such low resource translation direction are prone to hallucination. \citet{wang2023documentlevel, karpinska2023large} show that ChatGPT can match the performance of fully supervised models for document-level translation. \citet{bang2023multitask} find that when it comes to translation from high-resource languages into English, ChatGPT is comparable with the fully supervised model authors use but that performance degrades by almost $50$\% when translating from low-resource languages into English. \citet{huang2023languages} propose a prompting technique called cross-lingual-thought prompting (XLT) to improve cross-lingual performance for a wide range of tasks, including MT. Similarly, \citet{lu2023error} asks ChatGPT to correct its mistakes as a way to improve the model translation quality. To accurately translate attributive clauses from Japanese to Chinese, a pre-edit scheme is proposed in \citet{gu2023linguistically}, which improves accuracy of the translation by $\sim35$\%. \citet{lu2023chainofdictionary} proposes Chain-of-Dictionary (CoD) prompting to solve rare word translation issues. Prompting with CoD improves the performance of ChatGPT for both X-En and En-X language directions.

% %  moving it to Appendix as Prof asked
% \input{emnlp2023-latex/Sections/Lit Review/lit-rev-table}

\noindent\textbf{Arabic MT.} Arabic MT to date has primarily focused on two main themes: translating MSA and translation of Arabic dialects. 

\noindent\textbf{MSA MT.} The development of MSA MT systems has gone through various stages, including rule-based systems~\cite{bakr2008hybrid,mohamed2012transforming,salloum2013dialectal} and statistical MT~\cite{habash2009improving,salloum2011dialectal,ghoneim2013multiword}. There have also been efforts to employ neural machine translation (NMT) \cite{bahdanau2014neural} methods for MSA. For instance, several sentence-based Arabic to English NMT systems, trained on different datasets, have been presented in~\newcite{akeel2014ann}, \newcite{junczys2016neural},  \newcite{almahairi2016first}, \newcite{durrani2017qcri}, and \newcite{alrajeh2018recipe}. Furthermore, researchers have explored Arabic-related NMT systems for translating from languages other than English to MSA, including Chinese~\cite{aqlan2019arabic}, Turkish~\cite{el2019translating}, Japanese~\cite{noll2019simple}, four foreign languages\footnote{English, French, German, and Russian.} \cite{nagoudi2022_arat5}, and $20$ foreign languages\ \cite{nagoudi-2022-turjuman}.

\noindent\textbf{Dialectal Arabic MT.} A number of works focus on translating between  MSA and various Arabic dialects. For instance, both \newcite{zbib2012machine} and \cite{salloum2014sentence} combine MSA and dialectal data to build an MSA/dialect to English MT system. \newcite{sajjad2013translating} use MSA as a pivot language for translating Arabic dialects into English. \newcite{guellil2017neural} propose an NMT system for translating Algerian Arabic, written in a mixture of Arabizi and Arabic characters, into MSA. \newcite{baniata2018neural} present an NMT system for translating Levantine and Maghrebi dialects into MSA.\footnote{\textit{Levantine} includes Jordanian, Syrian, and Palestinian. \\  \textit{~~Maghrebi} covers Algerian, Moroccan, and Tunisian.} Furthermore, \newcite{sajjad2020arabench} introduce AraBench, an evaluation benchmark for dialectal Arabic to English MT, and evaluate several NMT systems under different settings such as fine-tuning, data augmentation, and back-translation.  To address  the challenge of unsupervised dialectal MT, both \newcite{farhan2020unsupervised} and \newcite{nagoudi-2021-Code-Mixed} propose a zero-shot dialectal NMT system, where the source dialect is not present in the training data. More recently, \newcite{nagoudi2022_arat5} employ Arabic text-to-text transformer (AraT5) models for translating from various Arabic dialects to English.\\

\noindent\textbf{ChatGPT for Arabic MT.}
\citet{khondaker2023gptaraeval} and \citet{alyafeai2023taqyim} evaluate ChatGPT for X-Arabic and Arabic-X translation pairs. \citet{khondaker2023gptaraeval} evaluate ChatGPT and other contemporary LLMs such as BloomZ~\cite{muennighoff2022crosslingual} in few-shot settings (0, 1, 3, 5, and 10) on four X-Arabic and two code-mixed Arabic-X language sets. They show that providing in-context examples to ChatGPT achieves comparable results to a supervised baseline. \citet{alyafeai2023taqyim} evaluate ChatGPT and GPT-4 on $4,000$ Arabic-English sentence pairs from \citet{ziemski2016united} and find a supervised SoTA model to outperform ChatGPT and GPT-4 by a significant margin. 
These works, however, only consider a limited number of Arabic varieties. They also do not conduct a thorough analysis of the LLMs for MT. Additionally, none of the works evaluate Bard. Our work bridges these gaps by performing a comprehensive evaluation of these systems on a wide range of Arabic varieties. We also conduct our study on novel in-house data that, to the best of our knowledge, is not presented in the training data of LLMs such as ChatGPT and Bard.  
% Our work is unique in evaluating on  a wider range of dialects and we offer a human study of Bard usefulness. We now introduce our datasets.
Other works have focused on evaluating smaller-sized Arabic language models~\cite{abu-farha-magdy-2021-benchmarking,inoue2021interplay,app12115720}, including on recent benchmarks~\cite{nagoudi2023dolphin, elmadany2022orca}. %% Add all related works.
% Arabic has a plethora of dialects, which is one reason that makes Arabic languages very complex

%% file: Sections/Lit_Review/lit-rev-table.tex
% \usepackage[normalem]{ulem}
\begin{table*}[]
\centering
% \footnotesize 
\setlength{\tabcolsep}{2pt}
\renewcommand{\arraystretch}{1.5} %to squeeze table content
\resizebox{!}{0.5\linewidth}{%
\begin{tabular}{p{3cm}p{2cm}p{3cm}p{4cm}p{2cm}p{4cm}p{4cm}}
\toprule
\textbf{Ref} & \multicolumn{1}{c}{\textbf{Focus}} & \multicolumn{1}{c}{\textbf{Languages}}       & \multicolumn{1}{c}{\textbf{Datasets}} & \multicolumn{1}{c}{\textbf{Setting}}     & \multicolumn{1}{c}{\textbf{Metrics}}     & \multicolumn{1}{c}{\textbf{Baselines}}                                 \\
\midrule
    \citet{jiao2023chatgpt} & Eval                           & Multi                               &              Flores-101, WMT-Bio/Rob                & ZS                              & BLEU                               & GoogleT, DeepL, Tencent                                             \\
    \citet{peng2023making} & Eval, Rob               & Multi                               &             Flores-200, WMT-News/Bio                 & ZS, FS                          & COMET, BLEU, ChrF                  & GoogleT                                                             \\
    \citet{gao2023design} & Eval, Prompting            & Multi/6TD                           &               Flores-101               & ZS, FS-1/5                    & BLUE, ChrF++, TER                  & GoogleT, DeepL                                                      \\
    \citet{zhu2023multilingual} & Eval                           & Multi(102)/202 TD                   &             Flores-101                 & ZS, FS                          & BLEU                               & XGLM-7.5B OPT-175B BLOOMZ-7.1B / SV-M2M-12B NLLB-1.3B \\
    \citet{hendy2023good} & Eval, Rob, DocLEval  & Multi(H, L)/18TD                    &                WMT-21/22              & ZS, FS-1/5                     & COMET, BLEU, ChrF, HE              & WMT-Best, MS-Translator                                            \\
    \citet{guerreiro2023hallucinations} & Eval, Hallucination            & Multi H, M, L / \textgreater 100 TD &             Flores, WMT, TICO                 & ZS                              & spBLEU, COMET, LaBSE               & SMaLL100, M2M                                                       \\
    \citet{wang2023documentlevel}& DocLEval                       & Multi H                             &              mZPRT, WMT-22, IWSLT-15/17, NewsComm-v11 Europar-v7,OpenSub-18                & ZS                              & BLEU, TER, COMET, dBLUE,T, HE      & MCN, GoogleT, MR-Doc2Doc, MR-Doc2Sent, Sent2Sent                    \\
    \citet{bang2023multitask}& Eval                     & Multi H, L 13/24 TD           & Flores-200             & ZS & ChrF++                             & FT-SOTA, ZS-SOTA                                                    \\
    \citet{huang2023languages}& Eval, Prompting         & Multi / 12 TD                       & FLORES                       &                                 & SacreBLEU                          & text-davinci-003                                                    \\
    % & Eval                           & Three / 2 TD                        & WMT-20                       & ZS, FS                          & BLEU, BERTScore, BLUER, COMET      & text-davinci-003                                                     \\
    \citet{gu2023linguistically}& Eval, Prompting         & Two /                               & NA                           & ZS                              & NA                                 & NA                                                                  \\
    \citet{karpinska2023large}& DocLEval                            & Multi/18 TD                         & Novel                        & ZS                              & COMET BLEURT BERTSCORE COMET-QE HE &                                                                     \\
    \citet{laskar2023systematic}& Eval                           & Multi/10TD                          & WMT14, WMT16, WMT19          & ZS                              & BLEU                               & PaLM-540B, Finetuned SOTA                                           \\
    \citet{ghosh2023chatgpt}& Eval, Fairness, Bias           & Multi / 5 TD                        & NA                           & ZS                              & HE                                 &                                                                     \\
    \citet{lu2023chainofdictionary}& Eval, Prompting                & Multi                               & Flores-200                   & ZS, FS-1/3                    & chrF++, BLEU                       & GPT-3.5-turbo                                                       \\
    \citet{ogundare2023comparative}& Eval                                & Multi                               & NA                           & ZS                              & SQ-Score     & GoogleT                                                             \\
    \citet{khondaker2023gptaraeval}& Eval                           & Multi/6 TD                          & UNPC, MDPC                   & ZS, FS-3/5/10                & BLUE                               & Supervised (AraT5)                                                  \\
    \citet{alyafeai2023taqyim}& Eval                           & Mono/1TD                            & UNv1                         & ZS, FS-3/5/10               & BLUE                               & Supervised SOTA                                                     \\
    \citet{Neubig_Zeno_GPT_Machine_2023}& Eval, Rob               & Multi                               & WMT                          & ZS, FS-1/5                    & COMET, ChrF,                        & GoogleT, MS Translate, DeepL       \\

\bottomrule
\end{tabular}
}\caption{\label{literature-review-table}
A summary of related works. We provide a brief description of recent studies aimed at evaluating LLMs on MT tasks. MT - machine translation. TD - translation direction. ZS - zero-shot, FS - few-shot, Rob - Robustness, H, L, M - high, low, medium resource.  %\textcolor{red}{Abdul2Abdul: Please rectify the abbreviation bro!}
% \moa {\hl{Please add a caption here}}
}

\end{table*}

%% file: Tables/stat_sent_dia.tex
\begin{table}[t]
\renewcommand{\arraystretch}{1.1}
    \centering
    \begin{tabular}{l c c c}
        \toprule
         \textbf{Variety} & \textbf{Mean} & \textbf{Median} & \textbf{Mode} \\
         \midrule
        CA & 22.98 & 19 & 15 \\
         MSA &  30.33 & 30 & 26\\ 
         \hdashline
         ALG & 15.63 & 13.5 & 10\\
         EGY & 19.42 & 16 & 13\\
         JOR & 15.50 & 14 & 11 \\
         MAU & 15.96 & 14 & 11 \\
         MOR & 17.63 & 17 & 17 \\
         
         PAL & 16.85 & 14.5 & 14 \\
         UAE & 14.98 & 13 & 10 \\
         YEM & 16.16 & 14 & 12 \\ \midrule
        \textbf{Avg.} & 18.52 & 16.45 & 13.9 \\
         \bottomrule
         
    \end{tabular}
    \caption{\label{tab:length_stats}
    Length statistics of the dataset (in number of words) across the different Arabic varieties.} 
    %\textcolor{red}{will come back to it! recomputation needed}}
\end{table}

%% file: Tables/results-tables/results-appendix.tex
\input{Tables/results-tables/table-caption}
% Justify pattern mismatch bw comet and bleu -? COMET is well suited for MSA because it is trained on MSA Arabic
\begin{table*}[h]
\centering
\footnotesize 
\renewcommand{\arraystretch}{1.4}   %to squeeze table content
\resizebox{1.0\linewidth}{!}{%
\begin{tabular}{llccccccccccccccc}
\toprule
                                   &                                  & \multicolumn{4}{c}{\textbf{ChatGPT}}                                                                                      & \multicolumn{2}{c}{\textbf{GPT-4}}                                  & \multicolumn{4}{c}{\textbf{Bard}}                                                                                         &                                                                                &                                                                                 &                                   &                                &                               \\
\multirow{-2}{*}{\textbf{Metrics}} & \multirow{-2}{*}{\textbf{Var/M}} & \textbf{0-shot}              & \textbf{1-shot}              & \textbf{3-shot}              & \textbf{5-shot}              & \textbf{0-shot}             & \textbf{5-shot}                       & \textbf{D1}                  & \textbf{D2}                  & \textbf{D2}                  & \textbf{Avg}                 & \multirow{-2}{*}{\textbf{\begin{tabular}[c]{@{}c@{}}NLLB\\ (SB)\end{tabular}}} & \multirow{-2}{*}{\textbf{\begin{tabular}[c]{@{}c@{}}NLLB\\ (Dia)\end{tabular}}} & \multirow{-2}{*}{\textbf{Amazon}} & \multirow{-2}{*}{\textbf{MST}} & \multirow{-2}{*}{\textbf{GT}} \\
\midrule
                                   \multirow{11}{*}{\rotatebox{90}{ChrF}}          & CA                              & 39.99           & 40.18           & 40.00           & 40.14           & \textbf{40.32}   & 39.25           & 38.56       & 37.44       & 38.87       & 38.29        & 28.56                                                                         & -                                                                              & 36.35                            & 38.09                         & 39.14                        \\
                                  & MSA                             & 69.37           & 69.84           & 69.91           & 70.15           & 69.04            & 69.56           & 63.15       & 60.71       & 61.94       & 61.93        & 65.27                                                                         & -                                                                              & 71.04                            & 70.35                         & \textbf{80.18}               \\
                                  & ALG                             & 40.04           & 41.27           & 41.72           & 41.75           & 43.97            & \textbf{42.91}  & 31.93       & 26.31       & 30.53       & 29.59        & 25.31                                                                         & -                                                                              & 33.15                            & 37.54                         & 33.96                        \\
                                  & EGY                             & 46.46           & 46.97           & 47.66           & 47.67           & \textbf{47.80}   & 47.62           & 42.96       & 39.62       & 43.83       & 42.14        & 33.03                                                                         & 36.68                                                                          & 40.43                            & 43.00                         & 43.35                        \\
                                  & JOR                             & 50.36           & 50.27           & 50.50           & 49.97           & 50.30            & 49.96           & 49.02       & 44.14       & 47.48       & 46.88        & 34.58                                                                         & 41.43                                                                          & 45.22                            & 47.48                         & \textbf{52.40}               \\
                                  & MAU                             & 32.77           & 32.01           & 32.91           & 32.97           & \textbf{34.90}   & 34.38           & 18.49       & 11.68       & 13.36       & 14.51        & 21.74                                                                         & -                                                                              & 29.86                            & 30.60                         & 28.74                        \\
                                  & MOR                             & 48.20           & 49.25           & 49.44           & 49.90           & 53.02            & \textbf{53.60}  & 47.40       & 46.98       & 47.73       & 47.37        & 27.22                                                                         & 39.04                                                                          & 34.79                            & 35.50                         & 39.36                        \\
                                  & PAL                             & 53.28           & 52.20           & 53.48           & 53.48           & \textbf{54.15}   & 53.42           & 41.54       & 39.69       & 44.43       & 41.89        & 35.68                                                                         & 40.02                                                                          & 45.79                            & 48.80                         & 48.64                        \\
                                  & UAE                             & 46.54           & 46.78           & 46.83           & 47.99           & 48.31            & \textbf{49.37}  & 39.31       & 36.39       & 39.68       & 38.46        & 30.02                                                                         & -                                                                              & 38.13                            & 41.42                         & 40.06                        \\
                                  & YEM                             & 40.70           & 41.54           & 41.60           & \textbf{42.35}  & 37.64            & 41.30           & 24.28       & 19.93       & 20.31       & 21.51        & 31.52                                                                         & 34.8                                                                           & 36.99                            & 39.29                         & 38.32                        \\
                                  \cline{2-17}
                                  & \textbf{Avg}                    & 46.77           & 47.03           & 47.41           & 47.64           & 47.94            & \textbf{48.14}  & 39.66       & 36.29       & 38.82       & 38.26        & 33.29                                                                         & 38.39                                                                          & 41.18                            & 43.21                         & 44.42                        \\
% comment this if you we chose to put this in the main
\midrule
\multirow{11}{*}{\rotatebox{90}{ChrF++}}          & CA                              & 37.89           & 38.15           & 38.04           & 38.22           & \textbf{38.31}   & 37.32           & 37.03       & 35.74       & 37.30       & 36.69        & 27.34                                                                         & -                                                                              & 34.65                            & 36.22                         & 37.44                        \\
                                  & MSA                             & 67.47           & 67.99           & 68.05           & 68.29           & 67.01            & 67.57           & 60.84       & 58.32       & 59.65       & 59.60        & 63.42                                                                         & -                                                                              & 68.99                            & 68.54                         & \textbf{79.00}               \\
                                  & ALG                             & 38.77           & 40.03           & 40.41           & 40.47           & 42.93            & \textbf{41.61}  & 31.18       & 25.69       & 29.83       & 28.90        & 24.16                                                                         & -                                                                              & 31.30                            & 35.20                         & 32.42                        \\
                                  & EGY                             & 45.13           & 45.69           & 46.47           & \textbf{46.54}  & 46.30            & 46.33           & 42.08       & 38.83       & 42.85       & 41.25        & 31.46                                                                         & 32.25                                                                          & 38.96                            & 41.41                         & 41.96                        \\
                                  & JOR                             & 49.42           & 49.36           & 49.58           & 49.03           & 48.72            & 48.87           & 48.15       & 43.34       & 46.60       & 46.03        & 33.32                                                                         & 40.3                                                                           & 43.94                            & 45.69                         & \textbf{51.30}               \\
                                  & MAU                             & 31.27           & 30.35           & 31.44           & 31.33           & \textbf{33.39}   & 32.76           & 18.03       & 11.63       & 13.08       & 14.25        & 20.27                                                                         & -                                                                              & 28.05                            & 28.44                         & 27.05                        \\
                                  & MOR                             & 47.71           & 48.69           & 48.93           & 49.42           & 52.57            & \textbf{53.14}  & 47.31       & 46.71       & 47.54       & 47.19        & 26.32                                                                         & 38.65                                                                          & 34.00                            & 34.76                         & 38.57                        \\
                                  & PAL                             & 52.26           & 51.10           & 52.48           & 52.50           & \textbf{53.12}   & 52.31           & 40.51       & 38.56       & 43.33       & 40.80        & 34.36                                                                         & 38.88                                                                          & 44.33                            & 47.16                         & 47.23                        \\
                                  & UAE                             & 45.82           & 45.88           & 45.94           & 47.19           & 46.44            & \textbf{48.54}  & 38.81       & 35.90       & 39.02       & 37.91        & 29.16                                                                         & -                                                                              & 37.32                            & 40.21                         & 39.11                        \\
                                  & YEM                             & 39.33           & 40.25           & 40.34           & \textbf{41.13}  & 36.38            & 39.93           & 23.78       & 19.76       & 19.94       & 21.16        & 30.07                                                                         & 33.69                                                                          & 36.09                            & 37.88                         & 36.99                        \\ \cline{2-17}
                                  & \textbf{Avg}                & 45.51           & 45.75           & 46.17           & 46.41           & 46.52            & \textbf{46.84}  & 38.77       & 35.45       & 37.91       & 37.38        & 31.99                                                                         & 37.35                                                                          & 39.76                            & 41.55                         & 43.11    \\
\midrule
 \multirow{11}{*}{\rotatebox{90}{TER}$\downarrow$}             & CA                              & 86.20           & 84.33           & 83.47           & 83.44           & 85.72            & 83.55           & 87.54       & 101.63      & 87.03       & 92.07        & 89.63                                                                         & -                                                                              & \textbf{81.83}                   & 83.86                         & 84.20                        \\
                                  & MSA                             & 44.73           & 43.56           & 43.19           & 42.70           & 44.13            & 43.77           & 55.07       & 67.96       & 62.54       & 61.86        & 44.79                                                                         & -                                                                              & 40.18                            & 39.52                         & \textbf{28.43}               \\
                                  & ALG                             & 87.08           & 80.86           & 80.25           & 78.48           & 80.56            & \textbf{78.91}  & 94.13       & 112.52      & 117.12      & 107.92       & 126.85                                                                        & -                                                                              & 90.62                            & 86.90                         & 89.43                        \\
                                  & EGY                             & 75.09           & 72.05           & 72.18           & \textbf{71.50}  & 73.44            & 71.61           & 75.22       & 81.33       & 77.04       & 77.86        & 88.69                                                                         & 86.29                                                                          & 80.56                            & 79.17                         & 76.40                        \\
                                  & JOR                             & 70.04           & 67.61           & \textbf{65.82}  & 67.10           & 70.35            & 68.46           & 68.07       & 73.85       & 69.41       & 70.44        & 108.25                                                                        & 80.83                                                                          & 72.71                            & 71.47                         & \textbf{65.82}               \\
                                  & MAU                             & 102.64          & 95.75           & 95.24           & 94.73           & 98.80            & 91.73           & 106.70      & 105.17      & 245.62      & 152.50       & 129.17                                                                        & -                                                                              & 96.85                            & \textbf{98.16}                & 99.54                        \\
                                  & MOR                             & 65.23           & 62.52           & 62.16           & 61.38           & 56.24            & \textbf{57.25}  & 61.44       & 61.89       & 61.25       & 61.53        & 100.23                                                                        & 73.39                                                                          & 82.60                            & 80.71                         & 77.75                        \\
                                  & PAL                             & 60.11           & 59.85           & 57.12           & 57.03           & \textbf{55.73}   & 57.38           & 73.29       & 75.46       & 66.10       & 71.62        & 86.76                                                                         & 78.23                                                                          & 67.38                            & 62.41                         & 65.84                        \\
                                  & UAE                             & 71.45           & 68.55           & 69.17           & 66.20           & 71.93            & \textbf{65.91}  & 79.58       & 76.24       & 73.60       & 76.47        & 85.07                                                                         & -                                                                              & 76.77                            & 73.87                         & 75.90                        \\
                                  & YEM                             & 84.96           & 82.09           & 80.51           & 81.45           & 85.53            & \textbf{80.81}  & 110.53      & 151.27      & 182.99      & 148.26       & 86.01                                                                         & 88.89                                                                          & 81.20                            & 80.58                         & 84.36                        \\
                                  \cline{2-17}
                                  & \textbf{Avg}                    & 74.75           & 71.72           & 70.91           & 70.40           & 72.24            & \textbf{69.94}  & 81.16       & 90.73       & 104.27      & 92.05        & 94.55                                                                         & 81.53                                                                          & 77.07                            & 75.67                         & 74.77                          \\
\bottomrule
\end{tabular}
}
\caption{\label{results-appendix}
Results in ChrF, ChrF++, and TER scores. Higher is better unless otherwise specified by $\downarrow$. Average represents the mean across all varieties. Three drafts (D1, D2, D3) from Bard are reported individually and averaged. NLLB is our MSA-based supervised baseline; NLLB (Dia) is dialect-specific. Abbreviations: SB - supervised baseline, Dia - dialect, Var - varieties, M - model, MST - Microsoft Translation, GT - Google Translate. Best results are in \textbf{bold}.
}
\end{table*}

%% file: Sections/robustness.tex
%%%%%%%%%% Muhammad: Comment bootsrap res and section
%%%%%%%%%%%%%%%%%%%%%%%%%%%%%
% \input{Tables/results-tables/diac_vs_no_diac_bootstrapped}
\subsection{Robustness of Results}\label{robustness-appendixx}
To more tightly ensure robustness of the results we acquire, we conduct bootstrap statistics with a maximum number of iterations of $1,000$ for BLEU, ChrF, ChrF++, and TER.\footnote{The bootstrapping process is quite compute-intensive. For example, to run the bootstrapping for the above mentioned four metrics, we parallelize the process over $48$ CPUs which takes over six hours to get all the results. While all metrics can be computed with CPU, COMET requires GPUs and running it over a similar amount of GPUs is not feasible. As a result of this constraint, we do not conduct bootstrapping for COMET.} 
Considering results of our bootstrapping experiment, we acquire results that are very close to those reported in Table~\ref{results-main}. For example, in our bootsrapping, the simple mean of means for all dialects is $23.69$ (std ±$2.85$) for ChatGPT (5-shot) compared to $23.64$ (std ±$2.73$) for GPT-4. In our results in Table (Table~\ref{results-main}) ChatGPT (5-shot) is $23.62$ compared to $23.64$ of GPT-4 (5-shot), in terms of BLEU score. We report the detailed results of bootstrapping in Table~\ref{results-main-bootstrapped}. 

% Adding this table just so that we can compare it with the main results. We have added this table to the appendix so please remove it from here before submission.

\input{Tables/results-tables/results-main-bootstrapped}

%% file: Tables/results-tables/results-main-bootstrapped.tex
\input{Tables/results-tables/table-caption}
% Justify pattern mismatch bw comet and bleu -? COMET is well suited for MSA because it is trained on MSA Arabic
\begin{table*}[h]
\centering
\small  % Adjust the text size
\renewcommand{\arraystretch}{1.3}   %to squeeze table content low value means squeeze more
\resizebox{1.\linewidth}{!}{% reduce the text size, lower means smaller text size

\begin{tabular}{l@{\hspace{1pt}}l@{\hspace{1pt}}c@{\hspace{3pt}}c@{\hspace{3pt}}c@{\hspace{3pt}}c@{\hspace{3pt}}c@{\hspace{3pt}}c@{\hspace{3pt}}c@{\hspace{2pt}}c@{\hspace{2pt}}c@{\hspace{2pt}}c@{\hspace{2pt}}c@{\hspace{3pt}}c@{\hspace{3pt}}c@{\hspace{3pt}}c@{\hspace{3pt}}c@{\hspace{3pt}}}

% \begin{tabular}{l@{\hspace{1pt}}l@{\hspace{4pt}}c@{\hspace{4pt}}c@{\hspace{4pt}}c@{\hspace{4pt}}c@{\hspace{4pt}}c@{\hspace{4pt}}c@{\hspace{4pt}}c@{\hspace{4pt}}c@{\hspace{4pt}}c@{\hspace{4pt}}c@{\hspace{4pt}}c@{\hspace{4pt}}c@{\hspace{4pt}}c@{\hspace{4pt}}c@{\hspace{4pt}}c}

\toprule
                                   &                                  & \multicolumn{4}{c}{\textbf{ChatGPT}}                                                                                      & \multicolumn{2}{c}{\textbf{GPT-4}}                                  & \multicolumn{4}{c}{\textbf{Bard}}                                                                                         &                                                                                &                                                                                 &                                   &                                &                               \\
\multirow{-2}{*}{\textbf{Met}} & \multirow{-2}{*}{\textbf{Var/M}} & \textbf{0-shot}              & \textbf{1-shot}              & \textbf{3-shot}              & \textbf{5-shot}              & \textbf{0-shot}             & \textbf{5-shot}                       & \textbf{D1}                  & \textbf{D2}                  & \textbf{D2}                  & \textbf{Avg}                 & \multirow{-2}{*}{\textbf{\begin{tabular}[c]{@{}c@{}}NLLB\\ (SB)\end{tabular}}} & \multirow{-2}{*}{\textbf{\begin{tabular}[c]{@{}c@{}}NLLB\\ (Dia)\end{tabular}}} & \multirow{-2}{*}{\textbf{Amazon}} & \multirow{-2}{*}{\textbf{MST}} & \multirow{-2}{*}{\textbf{GT}} \\
\midrule
\multirow{11}{*}{\rotatebox{90}{BLEU}}   & CA                     & 11.19\textsuperscript{±1.94}  & 12.08\textsuperscript{±1.94} & 12.21\textsuperscript{±2.06} & 12.48\textsuperscript{±2.07} & 11.76\textsuperscript{±1.85} & 11.41\textsuperscript{±1.83} & 12.30\textsuperscript{±2.02}   & 10.92\textsuperscript{±2.62}   & 12.35\textsuperscript{±2.14}   & 12.30\textsuperscript{±2.02}   & 7.13\textsuperscript{±1.55}                                                          & -                                                                     & 11.22\textsuperscript{±2.03}            & 11.99\textsuperscript{±2.10}         & \textbf{14.23\textsuperscript{±2.72}}        \\
                         & MSA                    & 42.97\textsuperscript{±2.98}  & 44.08\textsuperscript{±3.13} & 44.32\textsuperscript{±3.05} & 44.84\textsuperscript{±3.16} & 42.94\textsuperscript{±3.11} & 43.54\textsuperscript{±2.76} & 36.38\textsuperscript{±3.58}   & 32.99\textsuperscript{±4.83}   & 34.97\textsuperscript{±5.01}   & 36.38\textsuperscript{±3.58}   & 41.38\textsuperscript{±3.75}                                                         & -                                                                     & 46.48\textsuperscript{±3.33}            & 47.23\textsuperscript{±3.48}         & \textbf{65.47\textsuperscript{±5.21}}        \\
                         & ALG                    & 14.54\textsuperscript{±2.57}  & 16.43\textsuperscript{±2.81} & 17.16\textsuperscript{±3.00} & 17.33\textsuperscript{±2.75} & \textbf{18.54\textsuperscript{±2.77}} & 18.08\textsuperscript{±2.74} & 14.95\textsuperscript{±3.30}   & 11.75\textsuperscript{±3.42}   & 13.38\textsuperscript{±4.05}   & 14.95\textsuperscript{±3.30}   & 6.81\textsuperscript{±2.01}                                                          & -                                                                     & 9.89\textsuperscript{±2.26}             & 11.42\textsuperscript{±2.30}         & 11.72\textsuperscript{±2.07}        \\
                         & EGY                    & 19.80\textsuperscript{±2.54}  & 21.03\textsuperscript{±2.49} & 21.36\textsuperscript{±2.37} & \textbf{21.67\textsuperscript{±2.47}} & 20.99\textsuperscript{±2.58} & 21.43\textsuperscript{±2.78} & 21.17\textsuperscript{±2.91}   & 19.26\textsuperscript{±3.26}   & 20.81\textsuperscript{±3.32}   & 21.17\textsuperscript{±2.91}   & 10.62\textsuperscript{±2.35}                                                         & 12.46\textsuperscript{±2.01}                                                          & 14.78\textsuperscript{±2.21}            & 16.62\textsuperscript{±2.52}         & 17.89\textsuperscript{±2.55}        \\
                         & JOR                    & 25.51\textsuperscript{±3.64}  & 26.59\textsuperscript{±3.65} & 27.43\textsuperscript{±3.67} & 26.90\textsuperscript{±3.60} & 24.56\textsuperscript{±3.04} & 25.25\textsuperscript{±3.10} & 26.97\textsuperscript{±3.51}   & 23.30\textsuperscript{±3.34}   & 25.08\textsuperscript{±3.16}   & 26.97\textsuperscript{±3.51}   & 12.93\textsuperscript{±3.80}                                                         & 18.31\textsuperscript{±2.84}                                                          & 21.13\textsuperscript{±3.22}            & 21.39\textsuperscript{±3.02}         & \textbf{29.55\textsuperscript{±4.06}}        \\
                         & MAU                    & 8.53\textsuperscript{±1.73}   & 8.93\textsuperscript{±1.87}  & 9.17\textsuperscript{±1.89}  & 8.96\textsuperscript{±2.00}  & 9.19\textsuperscript{±1.79}  & \textbf{9.96\textsuperscript{±1.97}}  & 5.72\textsuperscript{±1.71}    & 4.19\textsuperscript{±1.82}    & 2.64\textsuperscript{±1.45}    & 5.72\textsuperscript{±1.71}    & 3.37\textsuperscript{±1.65}                                                          & \multicolumn{1}{l}{}                                                  & 7.06\textsuperscript{±1.62}             & 6.79\textsuperscript{±1.65}          & 7.45\textsuperscript{±1.95}         \\
                         & MOR                    & 27.14\textsuperscript{±3.41}  & 28.12\textsuperscript{±3.50} & 28.87\textsuperscript{±3.18} & 29.81\textsuperscript{±3.32} & 32.86\textsuperscript{±3.38} & \textbf{33.40\textsuperscript{±3.46}} & 31.23\textsuperscript{±4.02}   & 30.52\textsuperscript{±3.83}   & 31.06\textsuperscript{±3.73}   & 31.23\textsuperscript{±4.02}   & 9.30\textsuperscript{±2.72}                                                          & 19.46\textsuperscript{±2.67}                                                          & 12.61\textsuperscript{±2.12}            & 14.25\textsuperscript{±2.15}         & 16.96\textsuperscript{±2.48}        \\
                         & PAL                    & 29.43\textsuperscript{±3.26}  & 29.37\textsuperscript{±3.00} & 31.46\textsuperscript{±3.24} & 31.42\textsuperscript{±3.27} & \textbf{31.81\textsuperscript{±3.00}} & 30.39\textsuperscript{±3.01} & 21.96\textsuperscript{±3.74}   & 20.21\textsuperscript{±3.77}   & 23.92\textsuperscript{±3.88}   & 21.96\textsuperscript{±3.74}   & 14.03\textsuperscript{±2.99}                                                         & 17.08\textsuperscript{±2.45}                                                          & 21.77\textsuperscript{±2.63}            & 24.08\textsuperscript{±2.71}         & 25.34\textsuperscript{±3.05}        \\
                         & UAE                    & 24.14\textsuperscript{±3.21}  & 24.52\textsuperscript{±3.09} & 24.49\textsuperscript{±3.38} & 26.00\textsuperscript{±3.52} & 23.92\textsuperscript{±3.17} & \textbf{26.84\textsuperscript{±3.31}} & 21.49\textsuperscript{±3.69}   & 19.30\textsuperscript{±3.34}   & 21.15\textsuperscript{±3.41}   & 21.49\textsuperscript{±3.69}   & 10.95\textsuperscript{±2.25}                                                         & \multicolumn{1}{l}{}                                                  & 16.65\textsuperscript{±2.51}            & 18.95\textsuperscript{±2.87}         & 19.36\textsuperscript{±2.86}        \\
                         & YEM                    & 14.79\textsuperscript{±2.08}  & 16.02\textsuperscript{±2.21} & 16.94\textsuperscript{±2.41} & \textbf{17.46\textsuperscript{±2.36}} & 13.98\textsuperscript{±2.23} & 16.14\textsuperscript{±2.32} & 9.49\textsuperscript{±2.85}    & 7.22\textsuperscript{±3.17}    & 6.29\textsuperscript{±3.12}    & 9.49\textsuperscript{±2.85}    & 9.28\textsuperscript{±1.72}                                                          & 12.46\textsuperscript{±2.01}                                                          & 14.29\textsuperscript{±1.98}            & 14.19\textsuperscript{±2.02}         & 13.18\textsuperscript{±2.10}        \\ \cline{2-17}
                         & Avg                    & 21.80\textsuperscript{±2.74}  & 22.72\textsuperscript{±2.77} & 23.34\textsuperscript{±2.83} & \textbf{23.69\textsuperscript{±2.85}} & 23.05\textsuperscript{±2.69} & 23.64\textsuperscript{±2.73} & 20.17\textsuperscript{±3.13}   & 17.97\textsuperscript{±3.34}   & 19.16\textsuperscript{±3.33}   & 20.17\textsuperscript{±3.13}   & 12.58\textsuperscript{±2.48}                                                         & 15.95\textsuperscript{±2.40}                                                         & 17.59\textsuperscript{±2.39}            & 18.69\textsuperscript{±2.48}         & 22.12\textsuperscript{±2.90}        \\ \midrule
\multirow{11}{*}{\rotatebox{90}{ChrF}}   & CA                     & 39.96\textsuperscript{±1.65}  & 40.18\textsuperscript{±1.67} & 40.04\textsuperscript{±1.73} & 40.09\textsuperscript{±1.77} & \textbf{40.34\textsuperscript{±1.61}} & 39.28\textsuperscript{±1.59} & 38.64\textsuperscript{±2.01}   & 37.53\textsuperscript{±2.61}   & 38.88\textsuperscript{±1.96}   & 37.98\textsuperscript{±1.98}   & 28.61\textsuperscript{±2.44}                                                         & -                                                                     & 36.39\textsuperscript{±1.89}            & 38.24\textsuperscript{±1.89}         & 39.29\textsuperscript{±2.26}        \\
                         & MSA                    & 69.44\textsuperscript{±1.90}  & 69.85\textsuperscript{±1.95} & 69.94\textsuperscript{±1.91} & 70.22\textsuperscript{±1.89} & 68.99\textsuperscript{±1.91} & 69.60\textsuperscript{±1.79} & 63.19\textsuperscript{±3.60}   & 60.76\textsuperscript{±4.51}   & 62.13\textsuperscript{±4.14}   & 61.22\textsuperscript{±3.96}   & 65.30\textsuperscript{±2.59}                                                         & -                                                                     & 70.97\textsuperscript{±2.19}            & 70.30\textsuperscript{±2.24}         & \textbf{80.16\textsuperscript{±3.09}}        \\
                         & ALG                    & 40.10\textsuperscript{±2.30}  & 41.29\textsuperscript{±2.42} & 41.75\textsuperscript{±2.42} & 41.79\textsuperscript{±2.35} & \textbf{44.11\textsuperscript{±2.40}} & 43.11\textsuperscript{±2.39} & 32.02\textsuperscript{±4.91}   & 26.55\textsuperscript{±4.80}   & 31.16\textsuperscript{±5.28}   & 28.09\textsuperscript{±5.16}   & 25.46\textsuperscript{±2.69}                                                         & -                                                                     & 33.15\textsuperscript{±2.15}            & 37.55\textsuperscript{±2.21}         & 34.03\textsuperscript{±2.36}        \\
                         & EGY                    & 46.34\textsuperscript{±2.28}  & 46.92\textsuperscript{±2.18} & 47.60\textsuperscript{±2.16} & 47.51\textsuperscript{±2.26} & \textbf{47.62}\textsuperscript{±2.25} & 47.55\textsuperscript{±2.27} & 42.91\textsuperscript{±3.33}   & 39.53\textsuperscript{±4.11}   & 43.71\textsuperscript{±3.41}   & 40.92\textsuperscript{±3.38}   & 33.05\textsuperscript{±2.71}                                                         & 36.69\textsuperscript{±2.48}                                                          & 40.28\textsuperscript{±1.98}            & 42.99\textsuperscript{±2.18}         & 43.26\textsuperscript{±2.52}        \\
                         & JOR                    & 50.20\textsuperscript{±2.91}  & 50.11\textsuperscript{±2.90} & 50.51\textsuperscript{±2.98} & 50.04\textsuperscript{±2.76} & 50.25\textsuperscript{±2.60} & 49.87\textsuperscript{±2.58} & 49.09\textsuperscript{±3.39}   & 44.06\textsuperscript{±3.84}   & 47.50\textsuperscript{±3.10}   & 45.21\textsuperscript{±3.20}   & 34.64\textsuperscript{±3.55}                                                         & 41.40\textsuperscript{±2.66}                                                          & 45.16\textsuperscript{±2.80}            & 47.48\textsuperscript{±2.61}         & \textbf{52.51\textsuperscript{±3.31} }       \\
                         & MAU                    & 32.74\textsuperscript{±1.99}  & 31.99\textsuperscript{±2.03} & 32.87\textsuperscript{±2.01} & 32.97\textsuperscript{±2.05} & \textbf{34.95\textsuperscript{±2.10}} & 34.42\textsuperscript{±2.21} & 18.50\textsuperscript{±3.44}   & 11.86\textsuperscript{±3.52}   & 13.53\textsuperscript{±3.66}   & 12.42\textsuperscript{±3.59}   & 21.72\textsuperscript{±2.40}                                                         & \multicolumn{1}{l}{}                                                  & 29.76\textsuperscript{±1.82}            & 30.57\textsuperscript{±1.82}         & 28.74\textsuperscript{±2.12}        \\
                         & MOR                    & 48.29\textsuperscript{±2.61}  & 49.16\textsuperscript{±2.65} & 49.46\textsuperscript{±2.50} & 49.93\textsuperscript{±2.63} & 53.02\textsuperscript{±2.65} & \textbf{53.69\textsuperscript{±2.69}} & 47.44\textsuperscript{±4.42}   & 47.04\textsuperscript{±4.27}   & 47.82\textsuperscript{±4.11}   & 47.30\textsuperscript{±4.21}   & 27.26\textsuperscript{±2.94}                                                         & 39.01\textsuperscript{±2.30}                                                          & 34.74\textsuperscript{±2.15}            & 35.50\textsuperscript{±2.09}         & 39.35\textsuperscript{±2.33}        \\
                         & PAL                    & 53.25\textsuperscript{±2.30}  & 52.23\textsuperscript{±2.17} & 53.58\textsuperscript{±2.32} & 53.49\textsuperscript{±2.33} & \textbf{54.19\textsuperscript{±2.36}} & 53.48\textsuperscript{±2.29} & 41.45\textsuperscript{±4.95}   & 39.87\textsuperscript{±4.91}   & 44.19\textsuperscript{±4.56}   & 41.31\textsuperscript{±4.69}   & 35.75\textsuperscript{±3.28}                                                         & 39.94\textsuperscript{±2.58}                                                          & 45.94\textsuperscript{±2.13}            & 48.78\textsuperscript{±2.16}         & 48.65\textsuperscript{±2.63}        \\
                         & UAE                    & 46.48\textsuperscript{±2.65}  & 46.85\textsuperscript{±2.67} & 46.86\textsuperscript{±2.75} & 47.92\textsuperscript{±2.78} & 48.39\textsuperscript{±2.89} & \textbf{49.38\textsuperscript{±2.72}} & 39.47\textsuperscript{±4.66}   & 36.28\textsuperscript{±4.24}   & 39.72\textsuperscript{±4.35}   & 37.43\textsuperscript{±4.45}   & 29.98\textsuperscript{±2.35}                                                         & \multicolumn{1}{l}{}                                                  & 38.10\textsuperscript{±2.23}            & 41.41\textsuperscript{±2.61}         & 40.23\textsuperscript{±2.79}        \\
                         & YEM                    & 40.81\textsuperscript{±2.15}  & 41.67\textsuperscript{±2.25} & 41.59\textsuperscript{±2.43} & \textbf{42.53\textsuperscript{±2.22}} & 37.54\textsuperscript{±3.00} & 41.16\textsuperscript{±2.52} & 24.44\textsuperscript{±4.65}   & 20.17\textsuperscript{±4.66}   & 20.78\textsuperscript{±4.99}   & 20.37\textsuperscript{±4.88}   & 31.48\textsuperscript{±2.06}                                                         & 34.83\textsuperscript{±2.09}                                                          & 36.96\textsuperscript{±2.04}            & 39.27\textsuperscript{±2.08}         & 38.32\textsuperscript{±2.15}        \\ \cline{2-17}
                         & Avg                    & 46.76\textsuperscript{±2.27}  & 47.03\textsuperscript{±2.29} & 47.42\textsuperscript{±2.32} & 47.65\textsuperscript{±2.30} & 47.94\textsuperscript{±2.38} & \textbf{48.15\textsuperscript{±2.30}} & 39.72\textsuperscript{±3.94}   & 36.37\textsuperscript{±4.15}   & 38.94\textsuperscript{±3.96}   & 37.23\textsuperscript{±3.95}   & 33.33\textsuperscript{±2.70}                                                         & 38.37\textsuperscript{±2.42}                                                          & 41.15\textsuperscript{±2.14}            & 43.21\textsuperscript{±2.19}         & 44.45\textsuperscript{±2.56}        \\ \midrule
\multirow{11}{*}{\rotatebox{90}{ChrF++}} & CA                     & 37.85\textsuperscript{±1.66}  & 38.16\textsuperscript{±1.68} & 38.08\textsuperscript{±1.76} & 38.18\textsuperscript{±1.80} & \textbf{38.33\textsuperscript{±1.64}} & 37.37\textsuperscript{±1.62} & 37.10\textsuperscript{±2.03}   & 35.84\textsuperscript{±2.60}   & 37.31\textsuperscript{±1.98}   & 36.33\textsuperscript{±2.00}   & 27.41\textsuperscript{±2.32}                                                         & -                                                                     & 34.69\textsuperscript{±1.90}            & 36.37\textsuperscript{±1.92}         & 37.60\textsuperscript{±2.30}        \\
                         & MSA                    & 67.54\textsuperscript{±1.98}  & 68.01\textsuperscript{±2.03} & 68.08\textsuperscript{±2.00} & 68.35\textsuperscript{±1.99} & 66.96\textsuperscript{±2.01} & 67.61\textsuperscript{±1.84} & 60.88\textsuperscript{±3.52}   & 58.36\textsuperscript{±4.39}   & 59.84\textsuperscript{±4.05}   & 58.85\textsuperscript{±3.87}   & 63.45\textsuperscript{±2.66}                                                         & -                                                                     & 68.91\textsuperscript{±2.25}            & 68.49\textsuperscript{±2.32}         & \textbf{78.97\textsuperscript{±3.24}}        \\
                         & ALG                    & 38.84\textsuperscript{±2.32}  & 40.06\textsuperscript{±2.42} & 40.44\textsuperscript{±2.45} & 40.53\textsuperscript{±2.38} & \textbf{43.08\textsuperscript{±2.44}} & 41.80\textsuperscript{±2.42} & 31.25\textsuperscript{±4.77}   & 25.94\textsuperscript{±4.65}   & 30.44\textsuperscript{±5.08}   & 27.44\textsuperscript{±4.98}   & 24.34\textsuperscript{±2.59}                                                         & -                                                                     & 31.31\textsuperscript{±2.12}            & 35.22\textsuperscript{±2.22}         & 32.48\textsuperscript{±2.31}        \\
                         & EGY                    & 45.01\textsuperscript{±2.29}  & 45.67\textsuperscript{±2.20} & \textbf{46.40\textsuperscript{±2.15}} & 46.38\textsuperscript{±2.24} & 46.12\textsuperscript{±2.26} & 46.25\textsuperscript{±2.28} & 42.03\textsuperscript{±3.24}   & 38.76\textsuperscript{±3.99}   & 42.76\textsuperscript{±3.30}   & 40.09\textsuperscript{±3.28}   & 31.50\textsuperscript{±2.64}                                                         & 35.26\textsuperscript{±2.45}                                                          & 38.80\textsuperscript{±1.99}            & 41.40\textsuperscript{±2.17}         & 41.87\textsuperscript{±2.51}        \\
                         & JOR                    & 49.26\textsuperscript{±2.94}  & 49.20\textsuperscript{±2.93} & 49.59\textsuperscript{±3.01} & 49.08\textsuperscript{±2.79} & 48.68\textsuperscript{±2.59} & 48.80\textsuperscript{±2.60} & 48.21\textsuperscript{±3.36}   & 43.28\textsuperscript{±3.78}   & 46.62\textsuperscript{±3.10}   & 44.39\textsuperscript{±3.19}   & 33.39\textsuperscript{±3.52}                                                         & 40.27\textsuperscript{±2.69}                                                          & 43.87\textsuperscript{±2.78}            & 45.69\textsuperscript{±2.65}         & \textbf{51.40\textsuperscript{±3.35}}        \\
                         & MAU                    & 31.26\textsuperscript{±1.98}  & 30.35\textsuperscript{±2.00} & 31.40\textsuperscript{±2.00} & 31.32\textsuperscript{±2.04} & \textbf{33.44\textsuperscript{±2.05}} & 32.82\textsuperscript{±2.20} & 18.04\textsuperscript{±3.29}   & 11.80\textsuperscript{±3.35}   & 13.25\textsuperscript{±3.51}   & 12.28\textsuperscript{±3.44}   & 20.28\textsuperscript{±2.32}                                                         & -                                                                     & 27.94\textsuperscript{±1.80}            & 28.42\textsuperscript{±1.80}         & 27.07\textsuperscript{±2.07}        \\
                         & MOR                    & 47.79\textsuperscript{±2.65}  & 48.61\textsuperscript{±2.67} & 48.96\textsuperscript{±2.51} & 49.45\textsuperscript{±2.64} & 52.57\textsuperscript{±2.66} & \textbf{53.23\textsuperscript{±2.73}} & 47.36\textsuperscript{±4.36}   & 46.76\textsuperscript{±4.20}   & 47.64\textsuperscript{±4.05}   & 47.05\textsuperscript{±4.15}   & 26.40\textsuperscript{±2.96}                                                         & 38.62\textsuperscript{±2.28}                                                          & 33.95\textsuperscript{±2.12}            & 34.76\textsuperscript{±2.06}         & 38.56\textsuperscript{±2.33}        \\
                         & PAL                    & 52.22\textsuperscript{±2.34}  & 51.14\textsuperscript{±2.20} & 52.55\textsuperscript{±2.36} & 52.49\textsuperscript{±2.37} & \textbf{53.17\textsuperscript{±2.37}} & 52.38\textsuperscript{±2.30} & 40.41\textsuperscript{±4.83}   & 38.75\textsuperscript{±4.79}   & 43.07\textsuperscript{±4.48}   & 40.19\textsuperscript{±4.60}   & 34.43\textsuperscript{±3.21}                                                         & 38.80\textsuperscript{±2.56}                                                          & 44.47\textsuperscript{±2.12}            & 47.14\textsuperscript{±2.17}         & 47.26\textsuperscript{±2.63}        \\
                         & UAE                    & 45.76\textsuperscript{±2.67}  & 45.95\textsuperscript{±2.69} & 45.98\textsuperscript{±2.76} & 47.12\textsuperscript{±2.81} & 46.49\textsuperscript{±2.85} & \textbf{48.55\textsuperscript{±2.75}} & 38.96\textsuperscript{±4.59}   & 35.77\textsuperscript{±4.19}   & 39.07\textsuperscript{±4.29}   & 36.87\textsuperscript{±4.39}   & 29.13\textsuperscript{±2.33}                                                         & -                                                                     & 37.29\textsuperscript{±2.22}            & 40.21\textsuperscript{±2.64}         & 39.28\textsuperscript{±2.81}        \\
                         & YEM                    & 39.48\textsuperscript{±2.11}  & 40.43\textsuperscript{±2.22} & 40.36\textsuperscript{±2.38} & \textbf{41.37\textsuperscript{±2.19}} & 36.31\textsuperscript{±2.93} & 39.80\textsuperscript{±2.48} & 23.96\textsuperscript{±4.49}   & 20.00\textsuperscript{±4.51}   & 20.40\textsuperscript{±4.82}   & 20.13\textsuperscript{±4.71}   & 30.04\textsuperscript{±2.00}                                                         & 33.72\textsuperscript{±2.04}                                                          & 36.04\textsuperscript{±2.00}            & 37.86\textsuperscript{±2.03}         & 36.98\textsuperscript{±2.11}        \\ \cline{2-17}
                         & Avg                    & 45.50\textsuperscript{±2.29}  & 45.76\textsuperscript{±2.30} & 46.18\textsuperscript{±2.34} & 46.43\textsuperscript{±2.33} & 46.52\textsuperscript{±2.38} & \textbf{46.86\textsuperscript{±2.32}} & 38.82\textsuperscript{±3.85}   & 35.53\textsuperscript{±4.04}   & 38.04\textsuperscript{±3.87}   & 36.37\textsuperscript{±3.86}   & 32.04\textsuperscript{±2.65}                                                         & 37.33 \textsuperscript{±2.40}                                                         & 39.73\textsuperscript{±2.13}            & 41.56\textsuperscript{±2.20}         & 43.15\textsuperscript{±2.57}        \\ \midrule
\multirow{11}{*}{\rotatebox{90}{TER}$\downarrow$}    & CA                     & 86.32\textsuperscript{±4.42}  & 84.28\textsuperscript{±4.35} & \textbf{83.39\textsuperscript{±4.32}} & 83.50\textsuperscript{±4.62} & 85.72\textsuperscript{±4.59} & 83.33\textsuperscript{±4.27} & 87.71\textsuperscript{±5.06}   & 101.91\textsuperscript{±34.76} & 87.24\textsuperscript{±4.91}   & 97.02\textsuperscript{±4.96}   & 89.41\textsuperscript{±10.14}                                                        & -                                                                     & 81.81\textsuperscript{±3.67}            & 83.50\textsuperscript{±4.10}         & 83.87\textsuperscript{±5.04}        \\
                         & MSA                    & 44.64\textsuperscript{±3.13}  & 43.62\textsuperscript{±3.16} & 43.17\textsuperscript{±3.22} & 42.63\textsuperscript{±3.27} & 44.30\textsuperscript{±3.13} & 43.71\textsuperscript{±2.91} & 55.05\textsuperscript{±8.72}   & 67.26\textsuperscript{±16.34}  & 62.66\textsuperscript{±16.63}  & 65.73\textsuperscript{±13.99}  & 44.86\textsuperscript{±3.52}                                                         & -                                                                     & 40.43\textsuperscript{±3.45}            & 39.55\textsuperscript{±3.32}         & \textbf{28.59\textsuperscript{±4.78}}        \\
                         & ALG                    & 87.28\textsuperscript{±6.42}  & 80.95\textsuperscript{±4.94} & 80.14\textsuperscript{±4.88} & \textbf{78.33\textsuperscript{±4.70}} & 80.53\textsuperscript{±5.11} & 78.60\textsuperscript{±5.18} & 94.21\textsuperscript{±12.82}  & 111.99\textsuperscript{±35.96} & 115.62\textsuperscript{±37.82} & 113.20\textsuperscript{±29.49} & 128.00\textsuperscript{±46.39}                                                       & -                                                                     & 90.41\textsuperscript{±5.12}            & 86.93\textsuperscript{±5.48}         & 89.59\textsuperscript{±4.20}        \\
                         & EGY                    & 75.13\textsuperscript{±3.93}  & 71.94\textsuperscript{±3.74} & 72.12\textsuperscript{±3.43} & \textbf{71.38\textsuperscript{±3.77}} & 73.60\textsuperscript{±4.30} & 71.60\textsuperscript{±4.41} & 75.37\textsuperscript{±8.19}   & 81.23\textsuperscript{±10.53}  & 77.70\textsuperscript{±14.48}  & 80.05\textsuperscript{±12.38}  & 88.40\textsuperscript{±20.57}                                                        & 86.04\textsuperscript{±10.58}                                                         & 80.63\textsuperscript{±3.74}            & 79.12\textsuperscript{±4.67}         & 76.45\textsuperscript{±3.89}        \\
                         & JOR                    & 70.36\textsuperscript{±5.24}  & 67.77\textsuperscript{±4.98} & 66.04\textsuperscript{±4.70} & 67.04\textsuperscript{±4.85} & 70.33\textsuperscript{±4.44} & 68.46\textsuperscript{±4.25} & 68.13\textsuperscript{±6.01}   & 73.84\textsuperscript{±5.44}   & 69.47\textsuperscript{±4.86}   & 72.38\textsuperscript{±5.24}   & 108.32\textsuperscript{±35.11}                                                       & 80.80\textsuperscript{±4.20}                                                          & 72.97\textsuperscript{±4.72}            & 71.53\textsuperscript{±4.71}         & \textbf{65.56\textsuperscript{±5.27}}        \\
                         & MAU                    & 102.56\textsuperscript{±5.74} & 95.72\textsuperscript{±4.50} & 95.50\textsuperscript{±4.98} & 94.98\textsuperscript{±4.79} & 99.08\textsuperscript{±4.85} & \textbf{91.61\textsuperscript{±4.34}} & 107.13\textsuperscript{±11.27} & 104.58\textsuperscript{±9.07}  & 246.24\textsuperscript{±88.06} & 151.80\textsuperscript{±62.46} & 130.19\textsuperscript{±35.99}                                                       & -                                                                     & 96.82\textsuperscript{±4.16}            & 98.30\textsuperscript{±4.63}         & 99.65\textsuperscript{±5.09}        \\
                         & MOR                    & 65.20\textsuperscript{±3.91}  & 62.67\textsuperscript{±3.78} & 62.00\textsuperscript{±3.51} & 61.47\textsuperscript{±3.82} & \textbf{56.30\textsuperscript{±3.76}} & 57.06\textsuperscript{±4.05} & 61.46\textsuperscript{±4.88}   & 61.78\textsuperscript{±4.81}   & 61.13\textsuperscript{±4.56}   & 61.56\textsuperscript{±4.67}   & 100.36\textsuperscript{±24.83}                                                       & 73.51\textsuperscript{±3.13}                                                          & 82.81\textsuperscript{±3.77}            & 80.81\textsuperscript{±3.62}         & 77.68\textsuperscript{±3.89}        \\
                         & PAL                    & 59.96\textsuperscript{±3.89}  & 59.88\textsuperscript{±3.31} & 57.09\textsuperscript{±3.46} & 57.12\textsuperscript{±3.40} & \textbf{55.71\textsuperscript{±3.59}} & 57.32\textsuperscript{±3.55} & 72.77\textsuperscript{±11.10}  & 75.25\textsuperscript{±11.72}  & 66.45\textsuperscript{±4.84}   & 72.32\textsuperscript{±6.93}   & 86.11\textsuperscript{±19.44}                                                        & 78.47\textsuperscript{±3.08}                                                          & 67.31\textsuperscript{±3.17}            & 62.50\textsuperscript{±3.42}         & 65.88\textsuperscript{±4.06}        \\
                         & UAE                    & 71.55\textsuperscript{±5.21}  & 68.55\textsuperscript{±4.31} & 69.23\textsuperscript{±4.59} & 66.19\textsuperscript{±4.42} & 71.65\textsuperscript{±4.50} & \textbf{65.94\textsuperscript{±4.31}} & 78.90\textsuperscript{±9.64}   & 76.62\textsuperscript{±5.74}   & 73.59\textsuperscript{±4.99}   & 75.61\textsuperscript{±6.54}   & 85.08\textsuperscript{±7.65}                                                         & -                                                                     & 76.88\textsuperscript{±3.93}            & 73.83\textsuperscript{±4.22}         & 75.62\textsuperscript{±4.37}        \\
                         & YEM                    & 83.06\textsuperscript{±4.04}  & 80.08\textsuperscript{±3.86} & 79.04\textsuperscript{±4.02} & \textbf{79.47\textsuperscript{±4.20}} & 85.50\textsuperscript{±4.17} & 80.89\textsuperscript{±4.24} & 110.69\textsuperscript{±38.27} & 153.21\textsuperscript{±74.25} & 182.04\textsuperscript{±86.90} & 162.82\textsuperscript{±70.69} & 86.00\textsuperscript{±4.62}                                                         & 88.80\textsuperscript{±3.52}                                                          & 81.22\textsuperscript{±3.48}            & 80.47\textsuperscript{±3.88}         & 84.17\textsuperscript{±4.04}        \\ \cline{2-17}
                         & Avg                    & 74.61\textsuperscript{±4.59}  & 71.55\textsuperscript{±4.09} & 70.77\textsuperscript{±4.11} & 70.21\textsuperscript{±4.18} & 72.27\textsuperscript{±4.24} & \textbf{69.85\textsuperscript{±4.15}} & 81.14\textsuperscript{±11.60}  & 90.77\textsuperscript{±20.86}  & 104.21\textsuperscript{±26.80} & 95.25\textsuperscript{±21.73}  & 94.67\textsuperscript{±20.83}                                                        & 81.52\textsuperscript{±4.90}                                                          & 77.13\textsuperscript{±3.92}            & 75.65\textsuperscript{±4.21}         & 74.71\textsuperscript{±4.46}     \\      
\bottomrule
\end{tabular}
}
\caption{\label{results-main-bootstrapped}Bootstraped results for BLEU, ChrF, ChrF++, and TER with standard deviation in superscript. Higher is better unless otherwise specified by $\downarrow$. Average represents the mean across all varieties. Three drafts (D1, D2, D3) from Bard are reported individually and averaged. NLLB is our MSA-based supervised baseline; NLLB (Dia) is dialect-specific. Abbreviations: SB - supervised baseline, Dia - dialect, Var - varieties, M - model, MST - Microsoft Translation, GT - Google Translate. Best results are in \textbf{bold}.
}
\end{table*}

%% file: Tables/results-tables/diac_vs_no_diac-appendix.tex
% \usepackage{multirow}
\begin{table}[h]
\centering
\footnotesize 
\renewcommand{\arraystretch}{1.1}   %to squeeze table content
\resizebox{1.0\linewidth}{!}{%

\begin{tabular}%{cccccccccc}
{@{\hspace{1mm}}l@{\hspace{1mm}}l@{\hspace{1mm}}c@{\hspace{1mm}}c@{\hspace{1mm}}c@{\hspace{1mm}}c@{\hspace{1mm}}c@{\hspace{1mm}}c@{\hspace{1mm}}c@{\hspace{1mm}}c@{\hspace{1mm}}c@{\hspace{1mm}}}
\toprule
\multirow{2}{*}{Met}   & \multirow{2}{*}{Mo/Var} & \multirow{2}{*}{CGPT} & \multirow{2}{*}{GPT-4} & \multicolumn{2}{c}{Bard} & \multirow{2}{*}{NLLB} & \multirow{2}{*}{Amazon} & \multirow{2}{*}{MST} & \multirow{2}{*}{GT} \\

                       &                         &                          &                        & D1          & Avg        &                       &                         &                     &                     \\
\midrule
\multirow{2}{*}{ChrF}                    & CA                      & \textbf{50.59}                    & 50.35                  & 46.99       & \textbf{47.54}      & \textbf{37.76}                 & \textbf{40.08}                   & \textbf{42.73}               & \textbf{48.58}               \\
                                         & CA*                     & 50.01                    & \textbf{50.49}                  & \textbf{47.49}       & 47.35      & 32.13                 & 39.53                   & \textbf{42.73}               & 46.95               \\
\midrule
\multirow{2}{*}{ChrF++}                  & CA                      & \textbf{49.23}                    & 48.99                  & 46.11       & 46.74      & \textbf{37.09}                 & \textbf{39.33}                   & \textbf{41.95}               & \textbf{47.68}               \\
                                         & CA*                     & 48.97                    & \textbf{49.25}                  & \textbf{47.02}       & \textbf{46.81}      & 31.71                 & 38.78                   & \textbf{41.95}               & 45.93               \\
\midrule                                         
\multicolumn{1}{l}{\multirow{2}{*}{TER $\downarrow$}} & CA                      & 69.98                    & 67.17                  & 69.14       & 69.61      & 77.95                 & 73.45                   & \textbf{66.23}               & \textbf{62.76}               \\
\multicolumn{1}{l}{}                     & CA*                     & \textbf{68.48}                   & \textbf{66.04}                  & \textbf{64.82}       & \textbf{65.63}      & \textbf{75.42}                 & \textbf{68.95}                   & \textbf{66.23}               & 64.92    \\       
\bottomrule
\end{tabular}
}

\caption{\label{diac-vs-no-diac-appendix}
The effect of diacritics on translation quality. CA* is without diacritics. Higher is better unless otherwise specified by $\downarrow$. The best results are in \textbf{bold}. 
}
\end{table}

%% file: Tables/results-tables/diac_vs_no_diac_bootstrapped.tex
% \usepackage{multirow}
\begin{table}[h]
\centering
\footnotesize 
\renewcommand{\arraystretch}{1.4}   %to squeeze table content
\resizebox{1.0\linewidth}{!}{%

\begin{tabular}%{cccccccccc}
{@{\hspace{1mm}}l@{\hspace{1mm}}l@{\hspace{1mm}}c@{\hspace{1mm}}c@{\hspace{1mm}}c@{\hspace{1mm}}c@{\hspace{1mm}}c@{\hspace{1mm}}c@{\hspace{1mm}}c@{\hspace{1mm}}c@{\hspace{1mm}}c@{\hspace{1mm}}}
\toprule
\multirow{2}{*}{Met}   & \multirow{2}{*}{Mo/Var} & \multirow{2}{*}{CGPT} & \multirow{2}{*}{GPT-4} & \multicolumn{2}{c}{Bard} & \multirow{2}{*}{NLLB} & \multirow{2}{*}{Amazon} & \multirow{2}{*}{MST} & \multirow{2}{*}{GT} \\

                       &                         &                          &                        & D1          & Avg        &                       &                         &                     &                     \\
\midrule
\multirow{2}{*}{\rotatebox{90}{BLEU}}                       & CA                      & 23.47 \textsuperscript{± 2.54}                                 & 23.87 \textsuperscript{± 2.11}                               & 22.98 \textsuperscript{± 2.00}           & 22.99 \textsuperscript{± 1.99}            & \textbf{15.92} \textsuperscript{± 1.91}                              & 17.41 \textsuperscript{± 2.28}                                & \textbf{20.10} \textsuperscript{± 2.04}                            & \textbf{26.48} \textsuperscript{± 2.70}                            \\
                                            & CA*                     & \textbf{23.49} \textsuperscript{± 2.49}                                 & \textbf{24.50} \textsuperscript{± 2.04}                               & \textbf{25.33} \textsuperscript{± 1.86}           & \textbf{24.22} \textsuperscript{± 1.99}            & 13.51 \textsuperscript{± 2.06}                              & \textbf{18.67} \textsuperscript{± 2.38}                                & 20.02 \textsuperscript{± 2.03}                            & 24.48 \textsuperscript{± 2.56}                            \\
\midrule
\multirow{2}{*}{\rotatebox{90}{ChrF}}                       & CA                      & \textbf{50.60} \textsuperscript{± 1.79}                                 & 50.43 \textsuperscript{± 1.83}                               & 46.99 \textsuperscript{± 1.69}           & \textbf{47.68} \textsuperscript{± 1.77}            & \textbf{37.74} \textsuperscript{± 1.68}                              & \textbf{40.11} \textsuperscript{± 1.94}                                & \textbf{42.76} \textsuperscript{± 1.76}                            & \textbf{48.61} \textsuperscript{± 2.10}                            \\
                                            & CA*                     & 50.07 \textsuperscript{± 1.87}                                 & \textbf{50.58} \textsuperscript{± 1.69}                               & \textbf{47.50} \textsuperscript{± 1.86}           & 47.04 \textsuperscript{± 1.76}            & 32.08 \textsuperscript{± 1.93}                              & 39.61 \textsuperscript{± 1.87}                                & 42.65 \textsuperscript{± 1.71}                            & 46.88 \textsuperscript{± 1.95}                            \\
\midrule
\multirow{2}{*}{\rotatebox{90}{ChrF++}}  & CA                      & \textbf{49.24} \textsuperscript{± 1.83}                                 & 49.06 \textsuperscript{± 1.87}                               & 46.11 \textsuperscript{± 1.71}           & \textbf{46.89} \textsuperscript{± 1.78}            & \textbf{37.06} \textsuperscript{± 1.73}                              & \textbf{39.36} \textsuperscript{± 1.92}                                & \textbf{41.99} \textsuperscript{± 1.76}                            & \textbf{47.71} \textsuperscript{± 2.13}                            \\
\multicolumn{1}{l}{}                        & CA*                     & 49.03 \textsuperscript{± 1.97}                                 & \textbf{49.34} \textsuperscript{± 1.72}                               & \textbf{47.04} \textsuperscript{± 1.81}           & 46.42 \textsuperscript{± 1.75}            & 31.65 \textsuperscript{± 1.96}                              & 38.85 \textsuperscript{± 1.90}                                & 41.88 \textsuperscript{± 1.71}                            & 45.85 \textsuperscript{± 1.99}                            \\ \midrule
\multirow{2}{*}{\rotatebox{90}{TER}$\downarrow$}     & CA                      & 70.00 \textsuperscript{± 3.30}                                 & 67.08 \textsuperscript{± 2.60}                               & 69.10 \textsuperscript{± 2.71}           & 70.39 \textsuperscript{± 3.04}            & 77.96 \textsuperscript{± 2.84}                              & 73.40 \textsuperscript{± 3.08}                                & \textbf{66.19} \textsuperscript{± 2.55}                            & \textbf{62.74} \textsuperscript{± 2.96}                            \\
\multicolumn{1}{l}{}                        & CA*                     & \textbf{68.48} \textsuperscript{± 3.48}                                 & \textbf{65.97} \textsuperscript{± 2.77}                               & \textbf{64.93} \textsuperscript{± 2.69}           & \textbf{66.20} \textsuperscript{± 2.80}            & \textbf{75.42} \textsuperscript{± 2.30}                              & \textbf{68.89} \textsuperscript{± 2.56}                                & \textbf{66.19} \textsuperscript{± 2.49}                            & 65.04 \textsuperscript{± 2.77}                              \\       
\bottomrule
\end{tabular}
}

\caption{\label{diac-vs-no-diac-appendix-bootstrap}
Bootstrapped scores in BLEU, ChrF, ChrF++, and TER. CA* is without diacritics. Higher is better unless otherwise specified by $\downarrow$.}

\end{table}

%% file: Tables/homographs.tex
\begin{table}[]
    \centering
    \begin{tabular}{rlrl}
    \toprule
        \textbf{MSA} & \textbf{English} & \textbf{MSA} & \textbf{English}\\
        \midrule
         \<كَتَبَ> & He wrote & \<كُتُبٌ> & Books \\
         \<قَسَّمَ> & He divided & \<قَسَمٌ> & Oath\\
         \<عَلَمٌ> & Flag & \<عِلْمٌ> & Science \\
         \<صِدْقٌ> & Sincerity & \<صَدَّقَ> & He believed \\
         \<وُلِدَ> & He was born & \<وَلَدٌ> & Boy \\
         \<ذُرَةٌ> & Corn & \<ذَرَّةٌ> & Atom \\
         \<مَدْرَسَةٌ > & School & \<مُدَرِّسَةٌ> & Teacher \\
         \<حَمَّامٌ> & Bathroom & \<حَمَامٌ> & Pigeons \\
         \<حِدَادٌ> & Mourning & \<حَدَّادٌ> & Blacksmith \\
         \<شَعْرٌ> & Hair & \<شِعْرٌ> & Poetry \\
         \<مَرْكَبَةٌ> & Vehicle & \<مُرَكَّبَةٌ> & Composite \\
         \<سُكْرٌ> & Drunkenness&  \<سُكَّرٌ> & Sugar \\
         \<نَجَمَ> & It resulted & \<نَجْمٌ> & Star \\
         \<رَجُلٌ> & Man & \<رِجْلٌ> & Foot \\
         \<بَشَرٌ> & Humans & \<بَشَّرَ> & He preached \\
         \<مَلِكٌ> & King & \<مُلْكٌ> & Possession \\
         \<جَدٌّ> & Grandfather&  \<جِدٌّ> & Seriousness \\
         \<جَمَلٌ> & Camel & \<جُمَلٌ> & Sentences \\
         \<حَكَمٌ> & Referee & \<حُكْمٌ> & Ruling \\
         \<سَمَكٌ> & Fish & \<سُمْكٌ> & Thickness \\
     \bottomrule
    \end{tabular}
    \caption{Heterophonic homographs used to test model sensitivity to diacritics.}
    \label{tab:homographs}
\end{table}

%% file: Sections/appendix/evaluation.tex
\section{Evaluation and Baselines} \label{appendix-evaluation}
\subsection{Evaluation Metrics}\label{appendix-evaluation-metrics}
% \textbf{Evaluation Metrics.} 

\noindent\textbf{\textit{BLUE}}~\cite{bleu}. BLEU is used to evaluate machine translation quality by comparing n-gram ($n=4$) overlap between machine-generated translations and human references. Higher scores indicate better translation quality.\\
% no including for this iteration
\noindent\textbf{\textit{COMET.}}~\cite{comet} Cross-lingual Opus METric measures translation quality through source-to-translation word-level alignment. Higher values indicate better quality. We use the default model\footnote{https://huggingface.co/Unbabel/wmt22-comet-da} which supports Arabic. However, based on our inspection, we find that Arabic data used to train the model is mostly MSA. Hence, the model may not be able to capture dialect-level nuances in the source text while computing the scores. \\
\noindent\textbf{\textit{ChrF and ChrF++}}~\cite{chrf}. Character n-gram F-score calculates the F-score of character n-grams in the machine translation compared to the reference translations, with higher scores denoting better quality. ChrF++ is an extension of ChrF where the word order is 2.\\
\textbf{\textit{TER}}~\cite{snover-etal-2006-study}. Translation Error Rate measures translation quality by counting edit operations between the machine and reference translations, providing a lower score for better quality.\\
We use huggingface's implementation of these metrics in \textit{evaluate}\footnote{https://github.com/huggingface/evaluate} package. We use all the default parameters unless otherwise specified above. While BLEU, ChrF, and TER rely mostly on direct comparisons of tokens or characters between the MT output and reference, COMET uses a model-based approach to capture more complex aspects of the translation such as semantics.

\subsection{Baselines}\label{appendix-baselines}
\noindent\textbf{\textit{Google Translate.}} In 2016, Google replaced their Statistical Machine Translation (SMT) system with Google Neural Machine Translation (GNMT) \citet{gnmt} featuring an LSTM with 8 encoder layers and 8 decoder ones with attention and residual connections. GNMT was trained on Google's internal datasets and it supports 133 languages. GNMT currently is powered by Transformers.\\
\noindent\textbf{\textit{Microsoft Translator.}} Microsoft's translation service uses an NMT model that supports $111$ different languages. \\
\noindent\textbf{\textit{Amazon Translation.}} Amazon Web Services (AWS) offer batch translation with their NMT models that can translate to and from $75$ languages.\\
\noindent\textbf{\textit{NLLB-200.}} No Language Left Behind \cite{nllbteam2022language} is an open-source Transformer model developed by META. 
%It uses Sparsely Gated Mixture of Experts (MoE) layers replacing the feed-forward layer in every 4th block. 
It was trained on FLORES-200 \cite{nllbteam2022language}, NLLB-MD \cite{nllbteam2022language}, and NLLB-Seed \cite{nllbteam2022language} for a total of 18B sentence pairs. It supports $202$ languages (and $40,000$ translation directions), $76$ of which are not supported by the aforementioned Google and Microsoft translation systems \citet{nllbteam2022language}. 

% \subsection{Error Analysis of ChatGPT and GPT-4:}\label{appendix-cgpt-error-analysis}
% Our error analysis of ChatGPT and GPT-4 on the Moroccan and Egyptian dialect shows some reoccurring error types spanning semantic, syntactic, grammatical and lexical dimensions. They are described as follows:

% \noindent\textbf{Idiomatic Errors:} The literal translation of multi-word expressions (MWEs), proverbs, and other region-specific language.

% \noindent\textbf{Code-Switching Errors:} Including both undetected code-switched words as well as dialectal words inaccurately treated as code-switched elements.

% \noindent\textbf{Named Entity Errors:} The literal translation of proper nouns or their substitution with different ones.

% \noindent\textbf{Tense Errors:} Alterations of verb tenses.

% \noindent\textbf{Addition/Omission Errors:} The presence of words in the target that don't appear in the source, as well as the leaving out of source words in the translation.

% \noindent\textbf{Polysemy Errors:} A source word with two or more related meanings being translated into the wrong meaning.

% \noindent\textbf{Homonymy Errors:} Translating from the homonym of the source word instead of the intended word itself.

% Examples of these errors are presented in Table \ref{tab:cgpt-error-analysis}.

% \input{Tables/CGPT-error-analysis}

\subsection{Human Analysis of Bard Helpfulness}\label{appendix-helpfulness}
\noindent

\input{Tables/bard-error-counts}

\begin{figure}[H]
    \centering
    \includegraphics[width = \columnwidth]{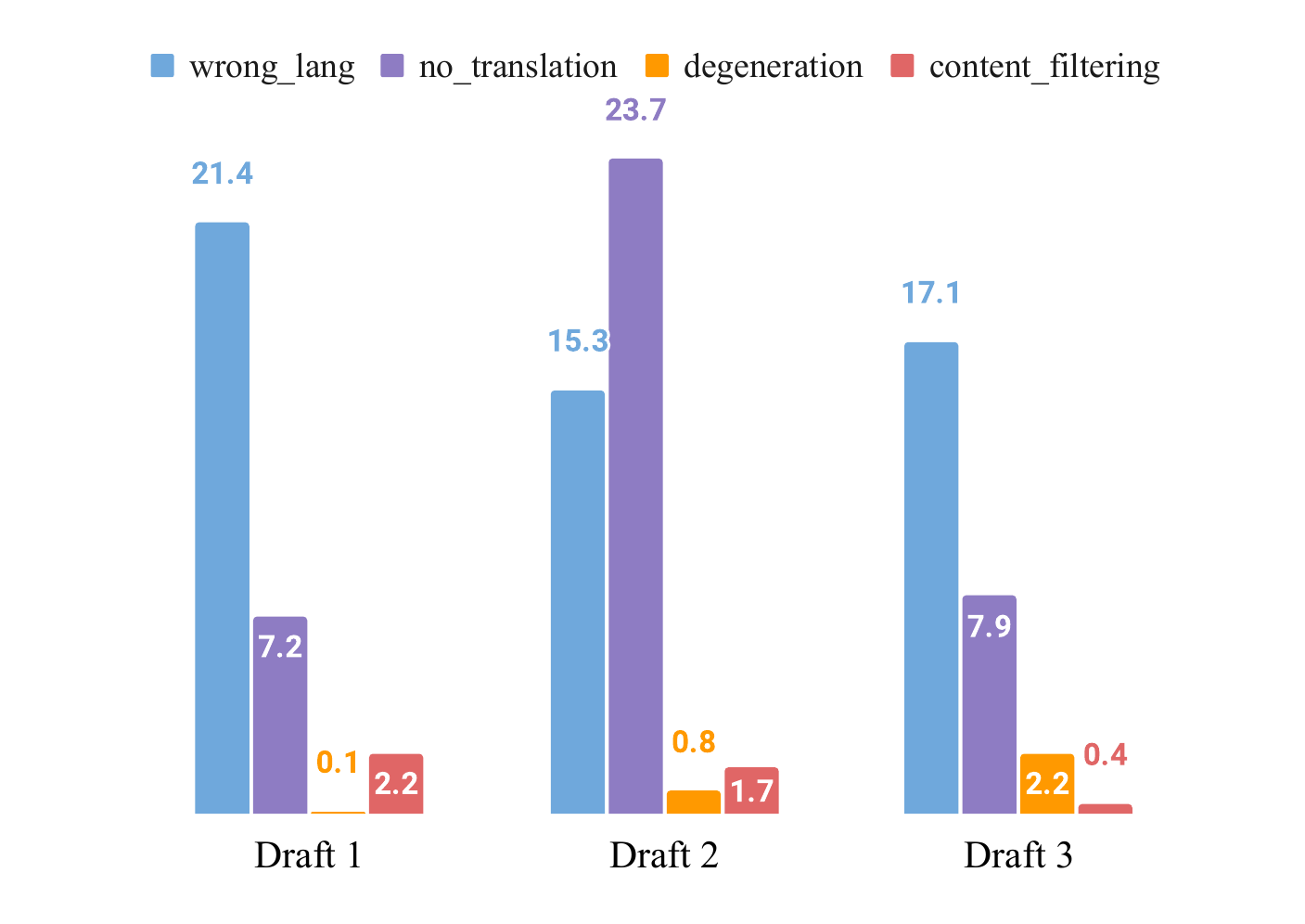}
    \caption{Percentage of Google Bard’s failure to follow the prompt for each draft relative to \textit{all errors across all drafts}.}
    \label{fig:drafts-chart}
\end{figure}

% \input{Tables/bard-fails}

% %%%%%%%%%%% Figure %%%%%%%%%%%%%%%%%%
\begin{figure*}
     \centering
     \begin{subfigure}{0.487\textwidth}
        \begin{framed}
             \centering
             \includegraphics[width=\textwidth]{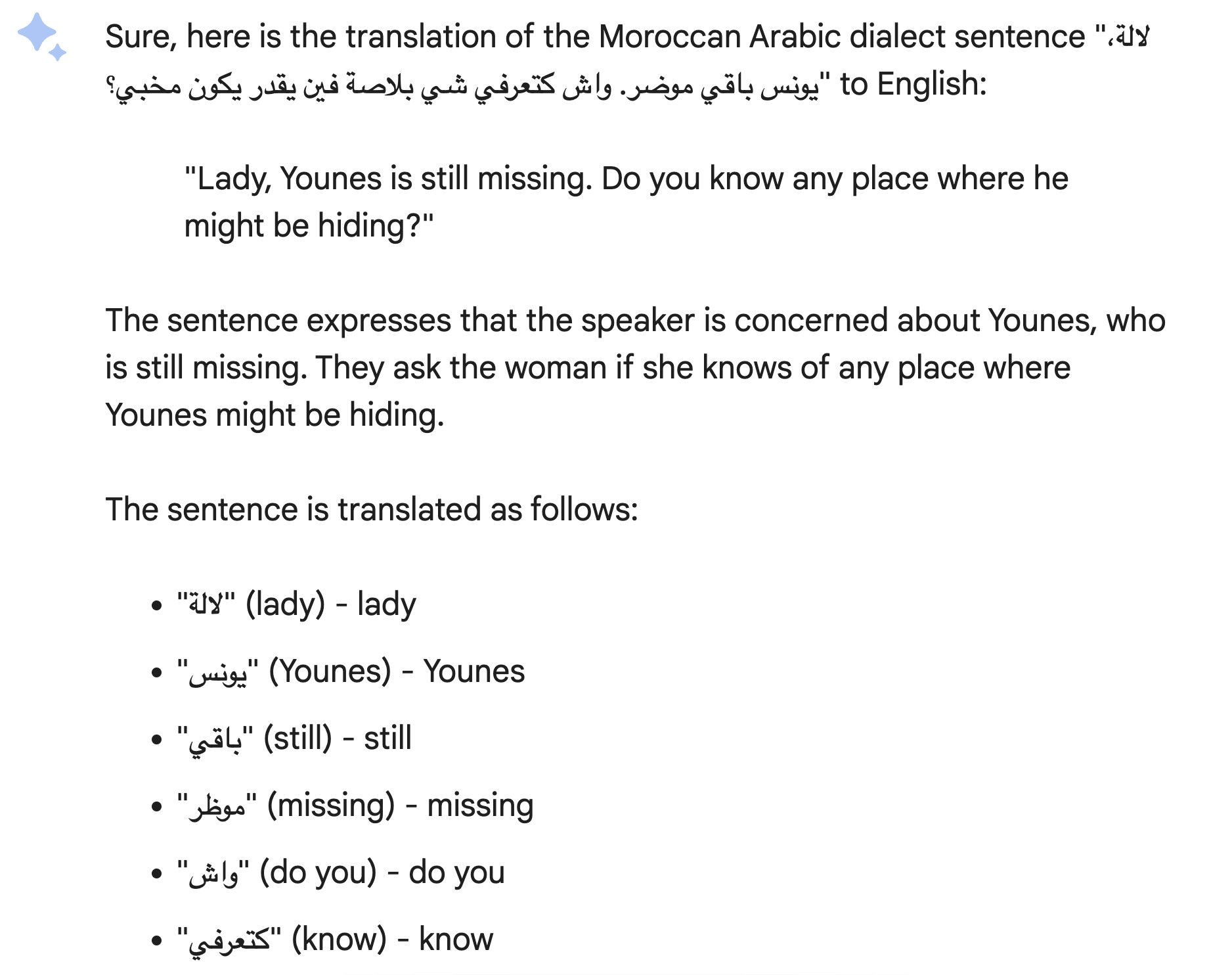}
             \end{framed}
           \caption{Google Bard's translation, explanation and breakdown of one dialectal sentence (from MOR).}
        \label{fig:bard-convo}
     \end{subfigure}
    \hfill
     \begin{subfigure}{0.468\textwidth}
        \begin{framed}
         \centering
         \includegraphics[width=\textwidth]{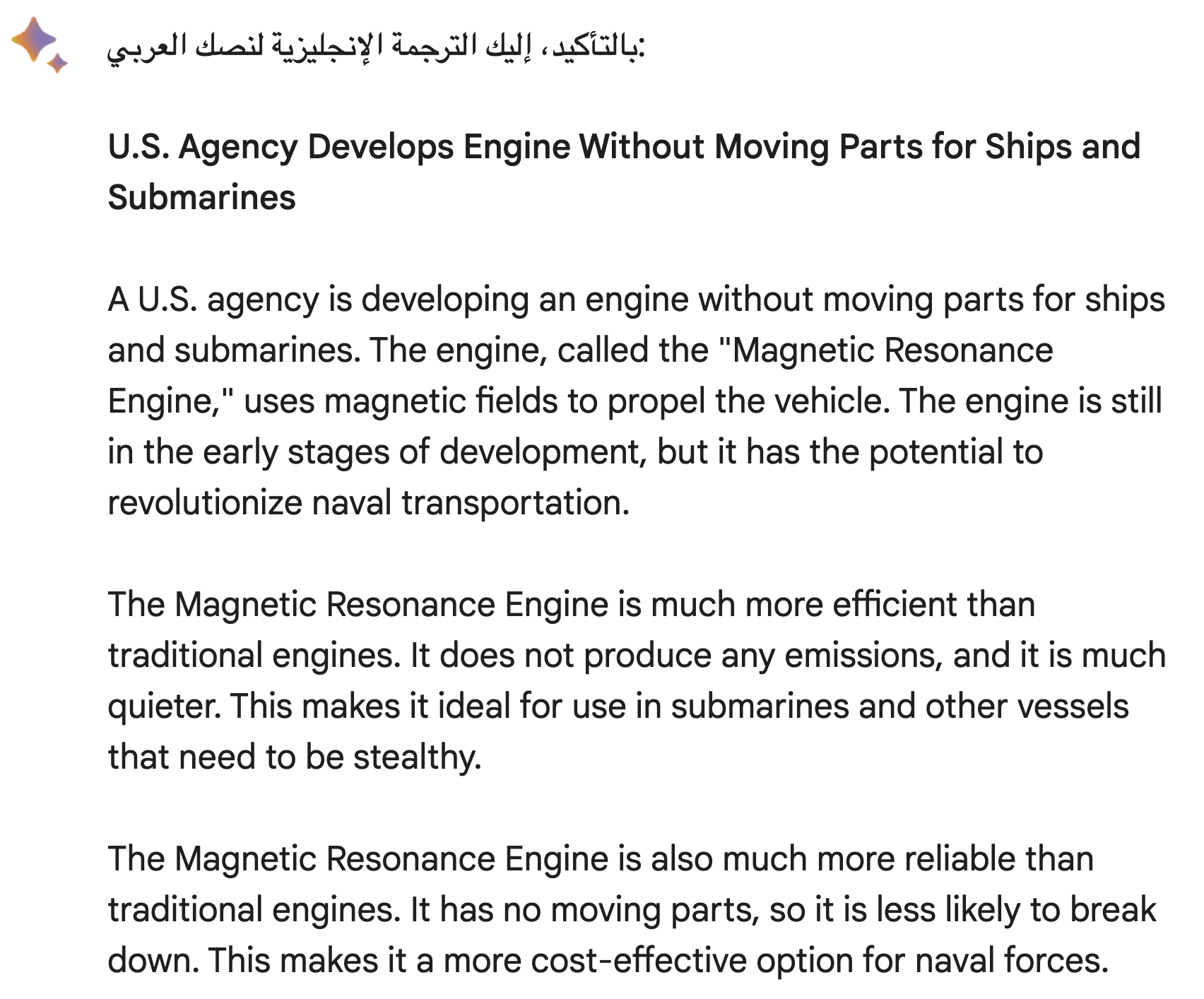}
         \end{framed}
         \caption{Google Bard's translation and context of an MSA sentence from the news domain. }
        \label{fig:bard-news}
     \end{subfigure}
     \caption{Examples of Google Bard's translation output. The bottom parts are cropped for readability.}
     \label{fig:bard-breakdown}
     \hfill
     
\end{figure*}

%% file: Tables/bard-error-counts.tex
\begin{table}
    \renewcommand{\arraystretch}{1.1}
    \centering
    \begin{tabular}{l c c c c c}
        \toprule
        \textbf{Variety} & \textbf{WL} & \textbf{NT} & \textbf{D} & \textbf{CF} & \textbf{Total} \\
        \midrule
        CA & 9 & 10 & 3 & 0 & 22 \\
        MSA & 1 & 27 & 1 & 0 & 29 \\
        \hdashline
        ALG & 81 & 72 & 5 & 6 & 164 \\
        EGY & 11 & 36 & 2 & 17 & 66 \\
        JOR & 14 & 22 & 2 & 2 & 40 \\
        MAU & 160 & 136 & 6 & 1 & 303 \\
        MOR & 42 & 13 & 3 & 2 & 60 \\
        PAL & 57 & 13 & 0 & 8 & 78 \\
        UAE & 33 & 49 & 2 & 10 & 94 \\
        YEM & 178 & 44 & 10 & 0 & 232 \\
        \midrule
        \textbf{Total} & 586 & 422 & 34 & 46 & 1088 \\
        \bottomrule
    \end{tabular}
    \caption{Counts of the helpfulness errors that Google Bard suffers from by Arabic variety and error category.}
    \label{tab:my_label}
\end{table}

%% file: Tables/mt-error.tex
\begin{table*}[]
\centering
\footnotesize 
\renewcommand{\arraystretch}{1.2}   %to squeeze table content
\resizebox{1.\linewidth}{!}{%
\begin{tabular}{p{0.5cm}p{6cm}p{5cm}p{8cm}}
% \rowcolor[HTML]{FFFFFF} 
\toprule
\textbf{Var} & \textbf{Source} & \textbf{Reference}  & \textbf{Translation} \\
\midrule
CA & \RaggedRight{\begin{arabtext} \small   فعلى العاقل أن يعلم أن الناس مشتركون مستوون في الحب لما يوافق، والبغض\end{arabtext}} & A wise man knows that people are the same in love and hate. & It is incumbent upon the wise person to know that people are equal in their love for what they agree upon, and in their hatred. - \textbf{ChatGPT} \\ 
& & & So it is upon the wise person to know that people are united and equal in their love for what is agreeable and in their hatred. - \textbf{GPT-4} \\
& & & So the wise man should know that people are united and equal in love for what agrees with them, and in hatred for what disagrees with them. - \textbf{Bard} \\
& & & A reasonable person would know that people share a common level of love for what is right and hate for what is wrong. - \textbf{NLLB} \\
& & & So the sane person should know that people are equal in love for what agrees and hate - \textbf{GT} \\
\midrule
MSA & \RaggedRight{\begin{arabtext} \small   تونس بين الصدمة والبحث عن حلّ لمحنة المهاجرين في صفاقس\end{arabtext}}  & Tunisia is between shock and looking for a solution to the Sfax migrant crisis. & Tunisia is between shock and searching for a solution to the plight of immigrants in Sfax. - \textbf{ChatGPT} \\
& & & Tunisia is between shock and the search for a solution to the migrant crisis in Sfax. \textbf{GPT-4}\\
& & & Tunisia between shock and the search for a solution to the migrant crisis in Sfax - \textbf{Bard} \\
& & & Tunisia is between shock and finding a solution to the migrant crisis in Safiq . \textbf{NLLB} \\
& & & Tunisia is between shock and the search for a solution to the plight of migrants in Sfax - \textbf{GT} \\

\midrule
ALG & \RaggedRight{\begin{arabtext} \small  لا نتا لا غيرك ما قادرين تحركو شعرة من راسي\end{arabtext}} & Neither you nor anyone else can scare me.. & You and no one else can move a single hair on my head. - \textbf{ChatGPT} \\

& & & Neither you nor anyone else can move a single hair on my head. - \textbf{GPT-4} \\
& & & Neither you nor anyone else can move a hair on my head - \textbf{Bard} \\
& & & Neither you nor Neta can move a hair of my head . - \textbf{NLLB} \\
& & & Neither you nor anyone else can move a hair from my head - \textbf{GT}\\

\midrule
MOR & \RaggedRight{\begin{arabtext} \small   عنده واحد الوشام هنايا. راه كيجي معاه واحد الكلب مرة مرة بالليل\end{arabtext}} & He has a tattoo here. He's accompanied by a dog from time to time at night. & He has a tattoo on his arm. He always comes with a dog sometimes at night. - \textbf{ChatGPT}\\
& & & He has a tattoo here. Sometimes he comes with a dog at night. - \textbf{GPT-4}\\
& & & He has a tattoo here. He comes with a dog sometimes at night. - \textbf{Bard}\\
& & & He has one tattoo here. He's got a dog once a night. - \textbf{NLLB}\\
& & & He has one tattoo Hanaya. He saw Keiji with one of the dogs once at night - \textbf{GT}\\
\midrule
MAU & \RaggedRight{\begin{arabtext} \small حيبللا كلمو يغير هو مزال ألا مصصم علنو ليهي يشتغل \end{arabtext}} & Hebila talked to him, but he persisted in working.  & They told him to change, but he is still stubborn and refuses to work. - \textbf{ChatGPT}\\
& & & God willing, talk to him to change; he is still insistent on working for him. - \textbf{GPT-4}\\
& & & He has changed a lot but he is still determined to work hard - \textbf{Bard}\\
& & & No matter how much he changes , he 's still not a public designer . - \textbf{NLLB}\\
& & & He doesn't want to talk to him about changing, he is still not a public designer, so he can work - \textbf{GT}\\
\midrule
UAE & \RaggedRight{\begin{arabtext} \small   أنا أبغي أفهم أنت ليش مب راضي تقنع أن أنا وأنت واحد ؟\end{arabtext}} & I want to understand why you are not convinced that you and I are the same person. & I want to understand why you're not convinced that you and I are one? - \textbf{ChatGPT}\\
& & & I want to understand why you're not convinced that you and I are one. - \textbf{GPT-4}\\
& & & I want to understand why you are not willing to be convinced that we are one - \textbf{Bard}\\
& & & I want to understand why you 're so happy to convince me that you and I are one ? - \textbf{NLLB}\\
& & & I want to understand why you are not satisfied with being convinced that you and I are one? - \textbf{GT}\\
\bottomrule
\end{tabular}
}
\caption{\label{output-samples}
Translations generated by the LLMs, the supervised baseline and the best performing commercial system (Google Translate). Translations from ChatGPT, GPT-4 and Bard were obtained under the zero-shot setting.
}
\end{table*}